\documentclass[preprint,12pt]{elsarticle}

\usepackage{amsmath,amsfonts}
\usepackage{algorithmic}
\usepackage{array}
\usepackage[tight,footnotesize]{subfigure}
\usepackage{textcomp}
\usepackage{stfloats}
\usepackage{url}
\usepackage{verbatim}
\usepackage{graphicx}
\def\BibTeX{{\rm B\kern-.05em{\sc i\kern-.025em b}\kern-.08em
    T\kern-.1667em\lower.7ex\hbox{E}\kern-.125emX}}
\usepackage{balance}
\usepackage{booktabs}
\usepackage{caption}
\usepackage{makecell}
\usepackage{adjustbox}
\usepackage{wrapfig}
\usepackage[linesnumbered,ruled]{algorithm2e}
\usepackage{amssymb}
\usepackage{lineno}
\usepackage{float}
\usepackage{algorithmic}
\usepackage[tight,footnotesize]{subfigure}
\usepackage{booktabs}
\usepackage{caption}
\usepackage{makecell}
\usepackage{adjustbox}
\usepackage{wrapfig}
\usepackage{xcolor}
\usepackage[linesnumbered,ruled]{algorithm2e}

\newcommand{\cmark}{\ding{51}}%
\newcommand{\xmark}{\ding{55}}%

\journal{Journal Name}

\begin{document}
\sloppy
\setlength{\parskip}{0pt}

\begin{frontmatter}

\title{Standardization of Psychiatric Diagnoses --- Role of Fine-tuned LLM Consortium and OpenAI-gpt-oss Reasoning LLM Enabled Decision Support System}

\author[label1]{Eranga Bandara}
\ead{cmedawer@odu.edu}
\author[label1]{Ross Gore}
\ead{rgore@odu.edu}
\author[label2]{Atmaram Yarlagadda}
\ead{atmaram.yarlagadda.civ@health.mil}
\author[label4]{Anita H.\ Clayton}
\ead{AHC8V@uvahealth.org}
\author[label3]{Preston Samuel}
\ead{preston.l.samuel.mil@health.mil}
\author[label1]{Christopher K.\ Rhea}
\ead{crhea@odu.edu}
\author[label1]{Sachin Shetty}
\ead{sshetty@odu.edu}


\address[label1]{Old Dominion University, Norfolk, VA, USA}
\address[label2]{McDonald Army Health Center, Newport News, VA, USA}
\address[label3]{Blanchfield Army Community Hospital, Fort Campbell, KY, USA}
\address[label4]{Department of Psychiatry and Neurobehavioral Sciences, \\ University of Virginia School of Medicine, Charlottesville, VA, USA}

\begin{abstract}

The diagnosis of most mental disorders, including psychiatric evaluations, primarily depends on dialogues between psychiatrists and patients. This subjective process can lead to variability in diagnoses across clinicians and patients, resulting in inconsistencies and challenges in achieving reliable outcomes. To address these issues and standardize psychiatric diagnoses, we propose a Fine-Tuned Large Language Model (LLM) Consortium and OpenAI-gpt-oss Reasoning LLM-enabled Decision Support System for the clinical diagnosis of mental disorders. Our approach leverages fine-tuned LLMs trained on conversational datasets involving psychiatrist–patient interactions focused on mental health conditions (e.g., depression). These models are capable of identifying mental disorders with high accuracy based on natural language input. The diagnostic predictions from individual models are aggregated through a consensus-based decision-making process, refined by the OpenAI-gpt-oss reasoning LLM. We propose a novel method for deploying LLM agents that orchestrate seamless communication between the LLM consortium and the reasoning LLM, ensuring transparency, reliability, and responsible AI across the entire diagnostic workflow. Each LLM in the consortium was fine-tuned using the Unsloth library on Google Colab’s Tesla GPUs. To optimize performance on consumer-grade hardware, we employed Low-Rank Adaptation with 4-bit quantization (QLoRA). Experimental results demonstrate the transformative potential of combining fine-tuned LLMs with a reasoning model to create a robust and highly accurate diagnostic system for mental health assessment. A prototype of the proposed platform, integrating three fine-tuned LLMs with the OpenAI-gpt-oss reasoning LLM, was developed in collaboration with the U.S. Army Medical Research Team in Norfolk, Virginia, USA. To the best of our knowledge, this work represents the first application of a fine-tuned LLM consortium integrated with a reasoning LLM for clinical mental health diagnosis—paving the way for next-generation AI-powered eHealth systems aimed at standardizing psychiatric diagnoses.

\end{abstract}


\end{frontmatter}

\section{Introduction}

The diagnosis of mental disorders remains one of the most complex and nuanced challenges in clinical medicine. Unlike many physical illnesses, which can be confirmed through objective tests such as blood panels, imaging, or genetic screening, psychiatric diagnoses rely heavily on subjective assessment~\cite{mdd-5k}. The current diagnostic process is predominantly based on conversations between psychiatrists and patients, guided by standardized criteria outlined in manuals such as the Diagnostic and Statistical Manual of Mental Disorders, Fifth Edition (DSM-5)~\cite{dsm-5}. During clinical interviews, psychiatrists assess the patient's reported symptoms, behavioral cues, emotional responses, and historical context to reach a diagnosis. However, this approach introduces significant variability and subjectivity. Diagnoses can differ between clinicians assessing the same patient, and even the same clinician may interpret similar symptoms differently across patients~\cite{metal-health-challanges-ai}. This variability can be attributed to differences in clinical experience, interpersonal dynamics, cultural context, and interpretation of symptom severity or duration~\cite{digital-mental-health-challenges}. Moreover, time constraints, communication barriers, and unconscious bias can further compromise diagnostic consistency. These challenges contribute to misdiagnosis, delayed treatment, and inconsistent care, ultimately affecting patient outcomes~\cite{metal-health-challanges}.

In recent years, advances in artificial intelligence, particularly  LLMs~\cite{llm, llama-4}, have demonstrated exceptional capabilities in natural language understanding and reasoning. These developments open new avenues for enhancing the diagnostic process in psychiatry by introducing data-driven, scalable, and interpretable decision support systems. In this context, we propose a Fine-Tuned LLM Consortium and OpenAI-gpt-oss Reasoning LLM~\cite{reasoning-llms, gpt-oss, proof-of-tbi, deep-stride} enabled decision support system designed to assist and augment the clinical diagnosis of mental disorders. The LLMs in the system are trained/fine-tuned on curated conversational datasets that simulate or replicate psychiatrist–patient interactions related to mental health evaluations, such as those for depression, anxiety, PTSD, and schizophrenia~\cite{CDSS, mental-health-ml}. By fine-tuning a set of LLMs on these datasets, the system learns to recognize patterns indicative of specific mental disorders. Predictions from these models are then aggregated through a consensus-based reasoning process, led by OpenAI-gpt-oss, a dedicated reasoning LLM that evaluates and refines the diagnostic output. The specialized orchestration LLM agent~\cite{llm-agents, agentic-ai, nurolense} coordinates communication between components, ensuring an end-to-end workflow that is accurate, interpretable, and aligned with DSM-5 diagnostic standards~\cite{dsm-5-traumatic-stress}. To enable deployment in resource-constrained environments, each model is fine-tuned using the Unsloth library on Google Colab GPUs~\cite{google-tpu} with Low-Rank Adaptation and 4-bit quantization (QLoRA)~\cite{qlora}, ensuring high performance with minimal computational overhead. The proposed approach aims not to replace clinicians but to support them with a robust, evidence-based, and transparent tool that improves diagnostic precision and consistency. The following are our main contributions of this research.

\begin{enumerate}
    \item Fine-tuning a consortium of LLMs to analyze conversational data and predict diagnoses of mental disorders.
    \item Incorporating the OpenAI-gpt-oss reasoning model to provide the final diagnosis based on the LLM consortium's predictions.
    \item Automating the end-to-end diagnostic decision-making process for mental disorders through an LLM consortium integrated with the OpenAI-gpt-oss model, orchestrated by LLM agents to ensure transparency, reliability, and adherence to Responsible AI principles.
    \item Implementing the prototype of the platform, integrating three LLMs with OpenAI-gpt-oss reasoning LLM, in collaboration with the U.S. Army Medical Research Team in Norfolk, Virginia, USA.
\end{enumerate}

The remainder of the paper is organized as follows: Section 2 introduces the core technologies that underpin the proposed AI-assisted psychiatric diagnostic platform. Section 3 details the overall system architecture, highlighting the integration of large language models and reasoning engines. Section 4 outlines the platform's core functionalities and operational workflow. Section 5 presents implementation details and evaluates the system’s performance across diagnostic tasks. Section 6 reviews related work and contextualizes our approach within the landscape of LLM-based medical diagnosis systems. Finally, Section 7 concludes the paper and discusses potential directions for future research and clinical deployment.

\section{Background}

This section provides a foundational overview of the core technologies underpinning the proposed AI-assisted psychiatric diagnostic platform. In particular, we highlight advancements in Large Language Models (LLMs), reasoning-capable LLMs, fine-tuning techniques, and the emerging paradigm of AI agents. 

\subsection{Large Language Models (LLMs)}

Large Language Models (LLMs) are advanced deep neural networks trained on extensive text corpora to comprehend, generate, and reason using natural language. They form the backbone of modern Natural Language Processing (NLP) systems~\cite{deep-learning-llm} and have demonstrated exceptional performance across a wide range of tasks, including text summarization, machine translation, dialogue generation, and question answering~\cite{llm}.

Several prominent LLMs such as OpenAI’s GPT~\cite{gpt-llm}, Meta’s Llama~\cite{llama-3, llama-4}, Mistral~\cite{mistral-llm}, and Alibaba’s Qwen2~\cite{qwen2} are available in both proprietary and open-source formats. Open-source LLMs offer significant advantages for healthcare applications, including transparency, customizability, and lower deployment costs. For instance, Llama-3\cite{llama-3, llama-4} is valued for its compact architecture and strong performance, even on resource-constrained systems. Mistral\cite{mistral-fine-tune} features optimized attention mechanisms for faster inference, while Qwen2~\cite{qwen2} is designed for multilingual and on-device use, making it suitable for scalable, privacy-preserving deployments.

\subsection{Reasoning LLMs}

While foundational LLMs excel in pattern recognition and natural language generation, they often lack the capacity for structured, multi-step reasoning. Reasoning LLMs\cite{reasoning-llms} address this limitation by being specifically designed or fine-tuned to synthesize diverse inputs, resolve conflicting information, and support logical decision-making processes. Unlike traditional LLMs that primarily rely on next-token prediction, reasoning models simulate higher-order cognitive functions akin to human deductive reasoning\cite{gpt-llm, bassa-llama}.

OpenAI-gpt-oss~\cite{gpt-oss, o3} is a reasoning LLM designed to perform advanced evaluative and comparative tasks across multiple inputs. Unlike traditional generative LLMs that focus on single-output prediction, OpenAI-gpt-oss is capable of synthesizing responses, resolving contradictions, and applying logical inference to arrive at consistent, well-reasoned conclusions. It excels in tasks involving multi-model output reconciliation, ranking, and consensus generation. These capabilities make it particularly suitable for applications that require structured reasoning, such as diagnostic decision support, content validation, and multi-agent coordination, where interpretability and reliability are critical.

\subsection{LLM Fine-tuning}

Fine-tuning is a technique for adapting pre-trained LLMs to specific downstream tasks or domains. It involves retraining the model on curated datasets that reflect the target domain's language, structure, and semantics, allowing the model to produce outputs more aligned with specialized applications such as clinical diagnostics~\cite{llm-finetune, vindsec-llams}.

To optimize for efficiency and scalability of fine-tuning, Low-Rank Adaptation (LoRA)\cite{lora} is commonly used. LoRA introduces trainable low-rank matrices into the transformer architecture, enabling task-specific adaptation while significantly reducing the number of trainable parameters. In resource-constrained environments, Quantized LoRA (QLoRA)\cite{qlora} provides an even more memory-efficient approach by quantizing model weights to 4-bit representations. QLoRA retains most of the performance benefits of full-precision fine-tuning while dramatically reducing memory and compute requirements. Together, these techniques make fine-tuning large models feasible on modest hardware, supporting wider adoption of LLMs in specialized domains.

Several open-source libraries facilitate efficient fine-tuning workflows for LLMs. Unsloth\cite{llamafactory-unsloth}, for example, enables high-speed, memory-efficient fine-tuning of models such as LLaMA, Mistral, and Qwen using LoRA and QLoRA techniques. It is optimized for both consumer-grade GPUs (e.g., NVIDIA RTX 3090) and cloud-based environments, including TPU-enabled platforms like Google Colab\cite{google-tpu}. Successful fine-tuning of large models generally requires GPUs with ample VRAM and compute capabilities. High-performance GPUs such as the NVIDIA A100 (40GB/80GB) and H100 are well-suited for large-scale training workloads, while more accessible GPUs like the NVIDIA RTX 3090/4090 and Tesla T4 provide sufficient resources for small to medium-scale fine-tuning and prototyping~\cite{a100-gpu}.

\subsection{AI Agents and Agentic AI}

AI agents are autonomous computational entities designed to perform complex tasks by interacting with data sources, machine learning models, and external APIs within dynamic or uncertain environments. When these agents are powered by LLMs, they are referred to as LLM agents, capable of interpreting natural language instructions, generating structured outputs, managing tasks, and coordinating actions across digital ecosystems~\cite{llm-agents, agentic-ai}.

Agentic AI extends this concept by organizing multiple LLM agents into collaborative, role-specialized systems that demonstrate advanced capabilities such as long-term planning, self-reflection, adaptive behavior, and multi-agent coordination~\cite{agentic-ai, slice-gpt}. These systems operate through agent hierarchies or workflows in which each agent performs a specific role, such as prompt engineering, retrieval, inference, evaluation, or integration. The modularity of agentic architectures enhances scalability, interpretability, and reusability, making them particularly suitable for domains requiring structured reasoning, task delegation, and reliable decision support.

\section{System Architecture}

\begin{figure}[H]
\centering{}
\includegraphics[width=4.5in]{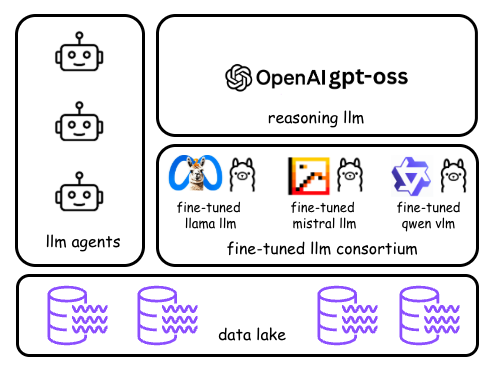}
\vspace{-0.1in}
\DeclareGraphicsExtensions.
\caption{Platform architecture.}
\label{indy528-architecture}
\end{figure}

Figure~\ref{indy528-architecture} describes the architecture of the platform. The proposed platform is composed of 4 layers: 1) Data Lake layer, 2) LLM Agent Layer, 3) LLM Layer, and 4) OpenAI-gpt-oss Reasoning Layer. Below is a brief description of each layer.

\subsection{Data Lake Layer}

The Data Lake layer serves as the foundational infrastructure for managing and storing extensive conversational datasets that simulate or replicate psychiatrist–patient interactions relevant to mental health evaluations. This centralized repository is designed to support the training and fine-tuning~\cite{wedagpt} of LLMs for predictive psychiatric diagnosis. It hosts a diverse collection of large-scale, labeled datasets comprising transcribed conversations, symptom narratives, and corresponding clinician-verified diagnoses—aligned with standardized diagnostic frameworks such as the DSM-5~\cite{dsm5-llm, dsm-5-criteria}. These datasets are critical for enabling the LLMs to learn clinically relevant patterns, understand nuanced symptom expressions, and associate them with specific mental disorders. By capturing a wide range of linguistic, contextual, and diagnostic variability, the Data Lake layer empowers the platform to develop robust, generalizable models capable of supporting accurate and consistent mental health assessments across diverse patient populations~\cite{me-llama, metavese-llama}.

\subsection{LLM Agent Layer}

The LLM agent layer functions as the orchestration and automation core of the platform, enabling seamless integration and coordination across the Data Lake, fine-tuned LLMs, and the OpenAI-gpt-oss reasoning engine. The LLM agents act as orchestrators, which are responsible for all custom prompt engineering required to facilitate effective communication between components, ensuring that diagnostic predictions are generated, aggregated, and refined in a coherent and structured manner. Specifically, the LLM agents dynamically construct prompts using patient-clinician conversational data retrieved from the Data Lake~\cite{prompt-engineering, llama-recipe}. These prompts are used to query the ensemble of fine-tuned LLMs, each of which outputs preliminary diagnostic assessments based on the detected symptoms and inferred clinical patterns. The agents then aggregate these outputs and format them into a consolidated, structured prompt tailored for the OpenAI-gpt-oss reasoning LLM~\cite{gpt-llm}. The OpenAI-gpt-oss model leverages its advanced reasoning capabilities to evaluate and synthesize the collective outputs of the LLM consortium, ultimately generating a refined and consistent diagnostic prediction. By adapting prompts to match the input requirements and context of each model, the LLM Agent Layer ensures optimal information flow and model interoperability~\cite{llm-agents}. This orchestrated process not only enhances diagnostic accuracy and consistency but also enables a fully automated, end-to-end AI-driven diagnostic workflow, as illustrated in Figure~\ref{llama2-flow}.

\subsection{LLM Layer}

\begin{figure}[H]
\centering{}
\includegraphics[width=5.2in]{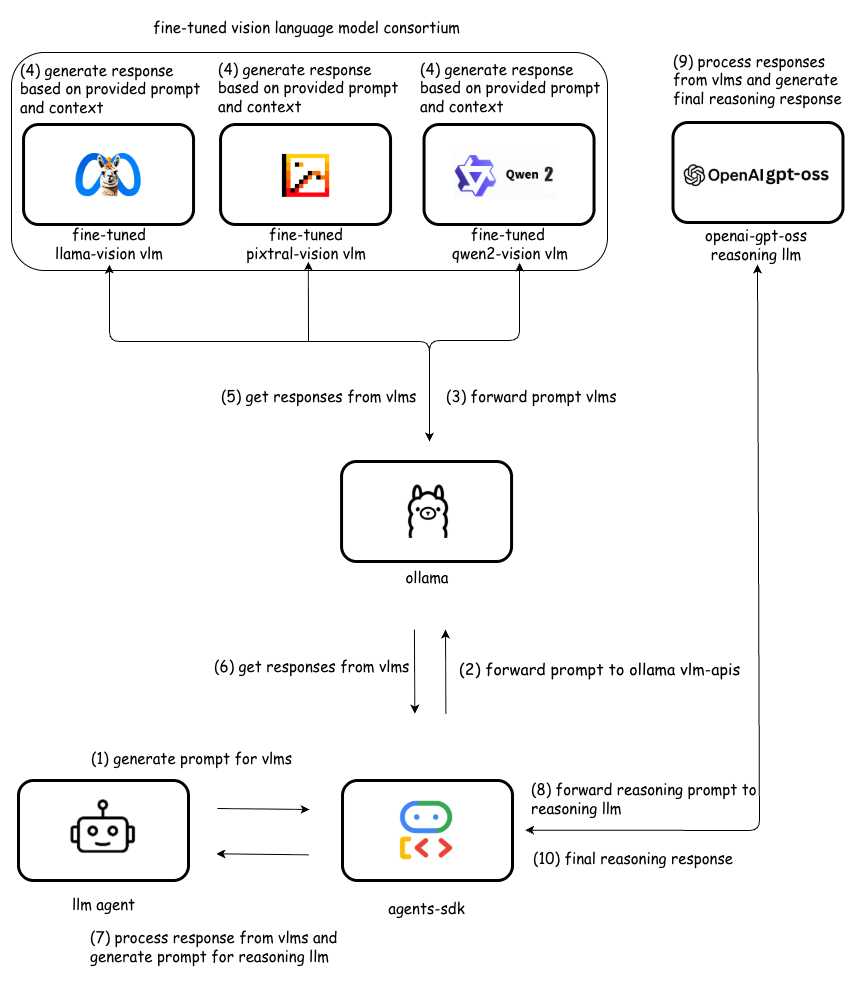}
\vspace{-0.1in}
\DeclareGraphicsExtensions.
\caption{LLM integration flow with Ollama LLM-API}
\label{llama2-flow}
\end{figure}

The LLM Layer serves as the analytical core of the platform, enabling the system to interpret psychiatrist–patient conversations and generate accurate diagnostic predictions. This layer comprises a consortium of fine-tuned LLMs, each trained on domain-specific conversational datasets involving mental health evaluations~\cite{llm-finetune}. These models are specialized to recognize linguistic and behavioral patterns indicative of various psychiatric conditions, such as depression, anxiety, PTSD, and schizophrenia~\cite{psychiatric-disorders}. The fine-tuned LLMs are deployed and managed using Ollama\cite{ollama, devsec-gpt}, a lightweight framework optimized for efficient inference and deployment of LLMs on consumer-grade hardware. This ensures the platform can scale effectively and maintain high performance even under resource constraints. As illustrated in Figure \ref{llama2-flow}, the LLM Agent Layer interfaces with the LLM consortium through Ollama's API, orchestrating prompt generation, model invocation, and response handling. By leveraging multiple specialized models within the consortium, the LLM Layer enhances diagnostic robustness through diversity in model reasoning, ultimately supporting a more comprehensive and consistent assessment of patient mental health based on natural language conversations.


\subsection{OpenAI-gpt-oss Reasoning LLM Layer}

The OpenAI-gpt-oss Reasoning Layer embodies the platform’s advanced cognitive and decision-making capabilities, leveraging state-of-the-art reasoning language models to synthesize diagnostic insights. The OpenAI-gpt-oss Reasoning LLM acts as the cognitive and synthesis engine of the platform. It is responsible for the high-level reasoning, integration, and refinement of system modeling predictions derived from the LLM consortium. 

Within the platform, OpenAI-gpt-oss serves as the final decision-making engine. It receives diagnostic predictions from the consortium of fine-tuned LLMs and performs structured reasoning to evaluate, cross-validate, and refine these outputs~\cite{llm-reasoning}. By synthesizing diverse model perspectives, OpenAI-gpt-oss determines the most consistent and clinically aligned diagnostic outcome, ensuring accuracy, coherence, and alignment with DSM-5 diagnostic criteria~\cite{dsm-5-criteria}. The LLM Agent Layer facilitates this process by aggregating the preliminary predictions and formatting them into structured, context-aware prompts tailored for OpenAI-gpt-oss. This enables the reasoning LLM to process heterogeneous inputs and deliver a final, consensus-driven diagnosis. By integrating probabilistic reasoning and consistency checks, the OpenAI-gpt-oss Reasoning Layer plays a pivotal role in enhancing the reliability, transparency, and clinical relevance of AI-assisted psychiatric diagnosis.

\section{Platform Functionality}

There are four main functionalities of the platform: 1) Data Lake Setup, 2) LLM Fine-Tuning, 3) Diagnosis prediction of Fine-tuned LLMs, and 4) Final Diagnosis Prediction by OpenAI-gpt-oss reasoning LLM. This section goes into the specifics of these functions.

\subsection{Data Lake Setup}

The first step in the platform’s workflow involves the setup of the Data Lake, which forms the foundational layer for storing, managing, and accessing large-scale conversational datasets. These datasets simulate or replicate psychiatrist–patient interactions focused on mental health evaluations and serve as the primary training resource for fine-tuning the language models used in diagnosis prediction~\cite{drhouse}. The Data Lake primarily contains transcribed conversations between clinicians and patients, along with their corresponding diagnoses provided by licensed psychiatrists. These diagnoses are annotated in accordance with standardized diagnostic frameworks, such as the DSM-5, ensuring clinical relevance and consistency~\cite{data-lake-llm}. This centralized repository enables the platform to efficiently organize diverse, high-quality training data necessary for modeling diagnostic reasoning and symptom interpretation. 

By providing robust, scalable, and secure data infrastructure, the Data Lake supports the development of fine-tuned LLMs capable of understanding complex psychiatric narratives, detecting subtle symptom patterns, and generating clinically accurate predictions. It is a critical enabler of the platform’s end-to-end AI-driven psychiatric diagnosis capabilities.

\subsection{LLM Fine-Tuning}

\begin{figure}[H]
\centering{}
\includegraphics[width=5.2in]{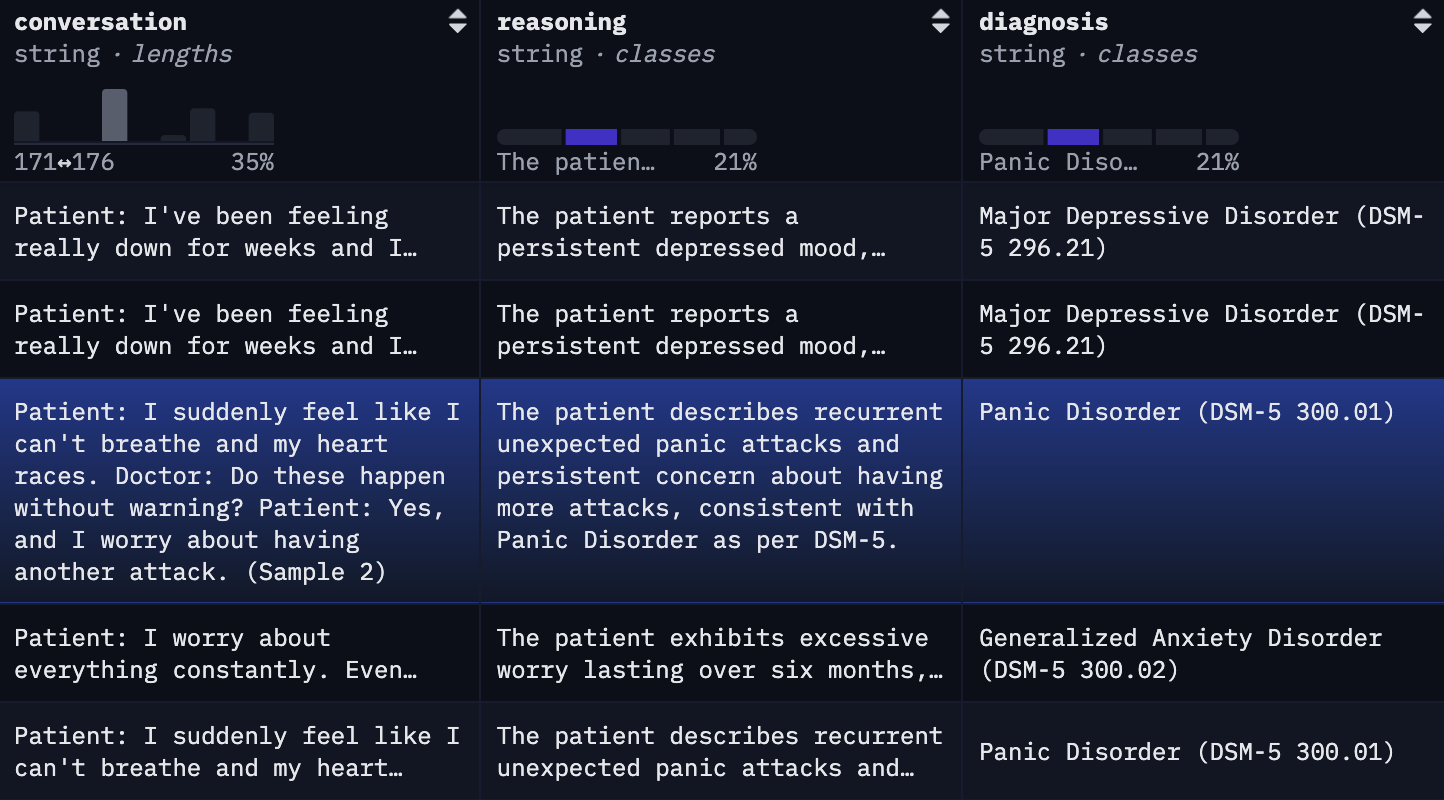}
\DeclareGraphicsExtensions.
\caption{The format of the dataset used to fine-tune the LLMs.}
\label{dataset-format}
\end{figure}

The second step in the platform workflow involves fine-tuning LLMs using the curated and pre-processed data stored in the Data Lake. This stage is crucial for transforming general-purpose models into specialized diagnostic agents capable of interpreting psychiatrist–patient conversations and identifying symptoms of mental disorders. Multiple state-of-the-art models, including Llama-3\cite{llama-3, llama-4}, Mistral\cite{mistral-fine-tune}, and Qwen2\cite{qwen2}, are fine-tuned on this domain-specific dataset to adapt them to the unique linguistic and contextual characteristics of psychiatric evaluations. The structure and composition of the dataset used for fine-tuning are illustrated in Figure\ref{dataset-format}. 

\begin{figure}[H]
\centering{}
\includegraphics[width=5.2in]{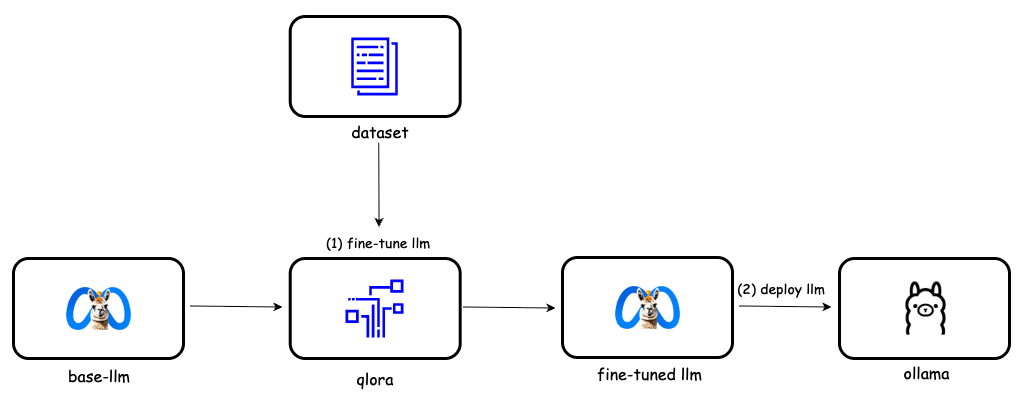}
\vspace{-0.1in}
\DeclareGraphicsExtensions.
\caption{Fine-tune LLMs with Qlora and deploy with Ollama.}
\label{llm-fine-tune}
\end{figure}

The fine-tuning process is carried out using the Unsloth library\cite{llamafactory-unsloth}, which facilitates efficient large-scale model adaptation. To ensure that models are deployable on consumer-grade hardware without compromising performance, the fine-tuning process incorporates Quantized Low-Rank Adapters (QLoRA)~\cite{qlora} with 4-bit quantization as depicted in Figure~\ref{llm-fine-tune}. This optimization significantly reduces memory and computational requirements, making the models suitable for real-time inference and edge deployment.

Upon completion, the fine-tuned and quantized models are deployed via Ollama~\cite{ollama, deep-stride}, a lightweight framework optimized to manage and run LLMs efficiently. These specialized models form the diagnostic core of the platform, each capable of analyzing psychiatric dialogue and producing preliminary mental disorder predictions based on learned diagnostic patterns and criteria aligned with DSM-5.

\subsection{Diagnosis Prediction by Fine-tuned LLMs}

Following the fine-tuning process, the next phase of the platform involves generating preliminary diagnoses of mental disorders using the consortium of fine-tuned LLMs. When new conversational data is received between a psychiatrist and a patient, the platform's LLM Agents initiate diagnostic analysis by interfacing with fine-tuned models through the Ollama API~\cite{ollama}. To facilitate accurate and context-aware predictions, the LLM Agent employs custom prompt engineering, embedding the relevant conversation data and structured diagnostic context into customized prompts for each model~\cite{prompt-engineering-rag}. These prompts are carefully designed to align with the LLM input expectations and to provide adequate clinical representation, such as duration of symptoms, emotional tone, and functional impact, based on the diagnostic criteria of DSM-5. Each fine-tuned model then analyzes the provided input, extracts potential symptoms, and produces its own diagnostic prediction. These individual outputs are collected by the LLM Agent, which organizes them into a structured format for downstream processing. This step ensures that the specialized capabilities of fine-tuned LLM are fully utilized to provide rich, reliable, and interpretable diagnostic insights into potential mental health conditions.

By allowing multiple independent evaluations throughout the model consortium, this layer improves diagnostic diversity, robustness, and the general ability of the platform to generalize in diverse patient presentations.

\subsection{Final Diagnosis Prediction by OpenAI-gpt-oss Reasoning LLM}

To ensure the highest level of diagnostic accuracy, reliability, and clinical validity, the platform employs a consensus-based decision-making mechanism to generate the final diagnosis. Rather than relying on the output of a single model, the platform aggregates diagnostic predictions from multiple fine-tuned LLMs within the consortium. These individual outputs are then evaluated, compared, and synthesized by OpenAI-gpt-oss, a specialized reasoning LLM designed to perform advanced analytical inference~\cite{llm-reasoning, nurolense}. As a core component of the architecture, OpenAI-gpt-oss acts as an intelligent adjudicator, capable of contextualizing, validating, and refining the predictions provided by the underlying LLMs. Using its advanced reasoning capabilities, OpenAI-gpt-oss identifies the most consistent and clinically appropriate diagnostic result from a diverse set of model-generated insights.

\begin{figure}[H]
\centering{}
\includegraphics[width=5.2in]{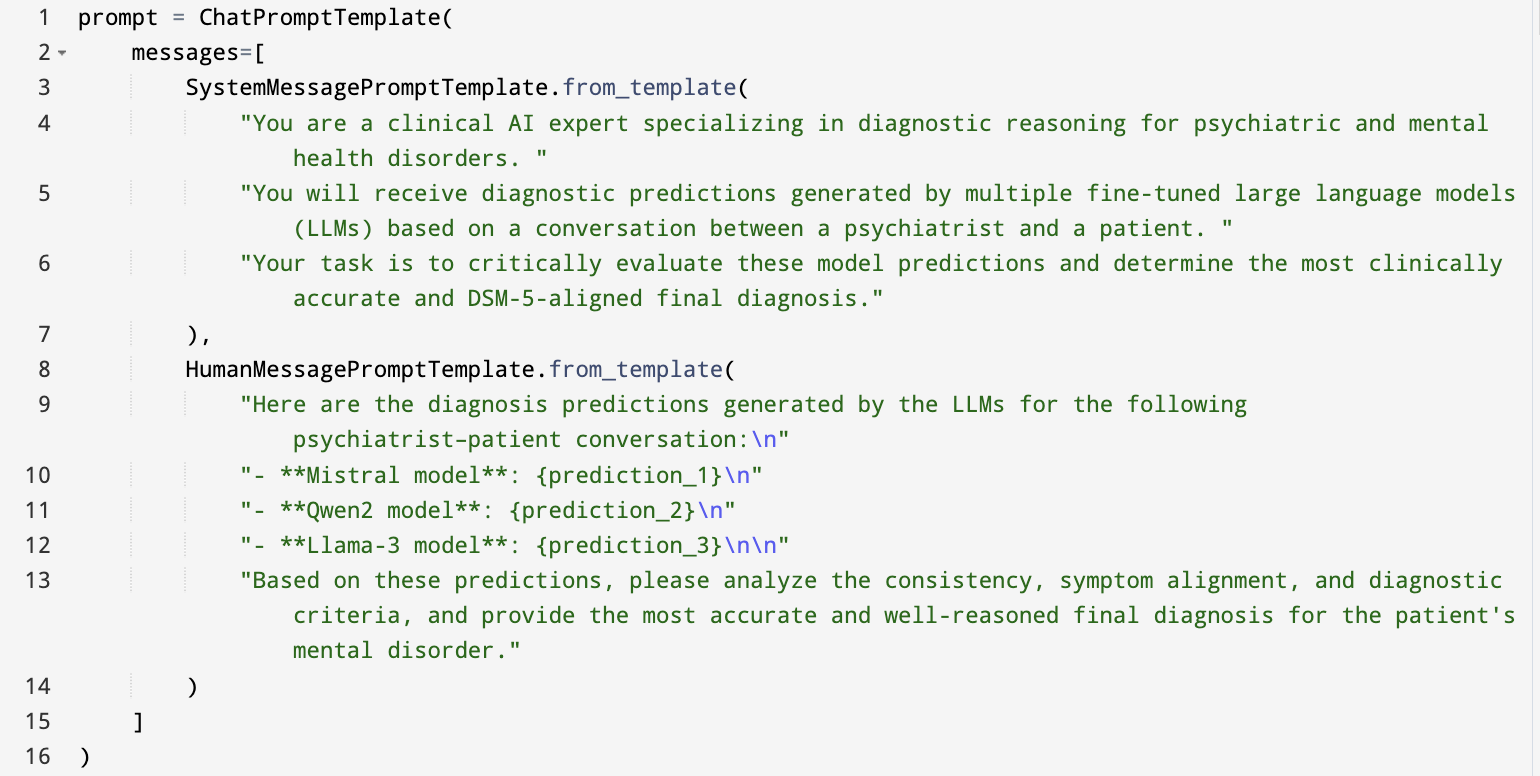}
\DeclareGraphicsExtensions.
\caption{Prompt for OpenAI-gpt-oss reasoning LLM for final prediction reasoning.}
\label{prompt}
\end{figure}

To enable this reasoning process, the LLM Agent constructs custom, structured prompts by embedding and organizing the outputs from the fine-tuned models. These prompts, as illustrated in Figure~\ref{prompt}, provide OpenAI-gpt-oss with a unified view of candidate diagnoses, associated symptoms, and contextual cues. The reasoning LLM processes this composite input and produces a final diagnosis that aligns with the DSM-5 criteria and reflects a well-supported clinical interpretation~\cite{dsm-5-criteria}. This consensus-driven architecture significantly improves the robustness and generalizability of diagnostic predictions by mitigating the limitations of individual models and reducing variability. By orchestrating this process through a transparent, explainable pipeline, the platform not only increases trustworthiness but also establishes a replicable framework for AI-assisted psychiatric evaluation.

The integration of ensemble-based inference with symbolic reasoning marks a transformative shift in mental health diagnostics, offering a scalable and interpretable decision support tool for clinicians. It demonstrates the potential of combining large-scale language understanding with structured reasoning to improve clinical decision-making in complex, subjective domains such as mental health.

\section{Implementation and Evaluation}

The implementation of the proposed platform was conducted in collaboration with the U.S. Army Medical Research team in Newport News, Virginia, USA. The LLM layer and reasoning Layer comprises of three fine-tuned LLMs, including Llama-3~\cite{llama-4}, Mistral~\cite{mistral-fine-tune}, and Qwen~\cite{qwen2}, and OpenAI-gpt-oss~\cite{reasoning-llms, o3} reasoning LLM. The LLM Agent Layer was implemented using OpenAI Agents SDK~\cite{openai-agent-sdk} and Google Agent Development Kit~\cite{agent-survey}, enabling secure orchestration, transparent auditability, and decentralized control of the LLM interactions. 


Fine-tuning was conducted using the Unsloth library~\cite{llamafactory-unsloth} on Google Colab, leveraging both NVIDIA A100 GPUs and Tesla TPUs~\cite{google-tpu} to support efficient and scalable training cycles. The original dataset consisted of approximately 2,000 annotated records, each containing a psychiatrist–patient conversation, the doctor’s diagnostic reasoning, and the corresponding final diagnosis. These records were compiled from multiple data sources, as illustrated in Figure~\ref{dataset-format}.

\begin{figure}[t]
\centering{}
\includegraphics[width=5.2in]{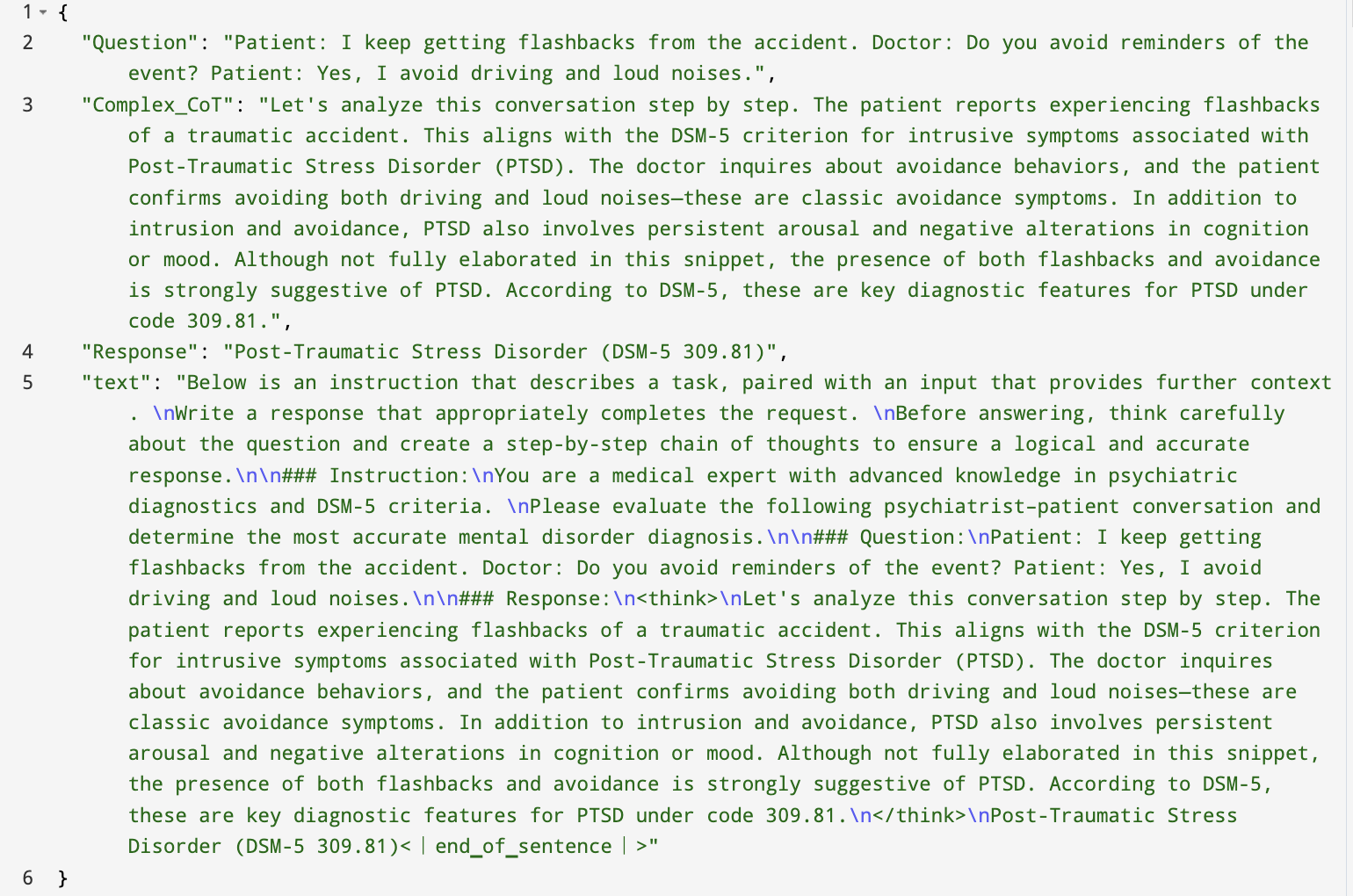}
\DeclareGraphicsExtensions.
\caption{The required data format of the unsloth library to fine-tune the LLM.}
\label{unsloth-format}
\end{figure}

The Unsloth framework requires the input data to be structured in a conversational format~\cite{llamafactory-unsloth}. To meet this requirement, the dataset was preprocessed and transformed into the required schema, shown in Figure~\ref{unsloth-format}. Each training sample includes the fields: instruction (providing context or a prompt to the LLM), content (representing the main conversational input), and text (containing the model's expected diagnostic reasoning output). The dataset was partitioned into training, validation, and testing subsets using a 2/3, 1/6, 1/6 split, respectively. The training process was completed in approximately 1,627 seconds (27.12 minutes). Peak memory reservation during training was 14.605 GB, with actual memory utilization reaching 5.853 GB, equivalent to 39.69\% of the reserved memory and 99.03\% of the peak allocation. These results demonstrate that fine-tuning large language models for psychiatric diagnosis using structured conversational data can be performed efficiently on moderate-scale datasets, even with limited hardware resources. This underscores the practicality and accessibility of applying LLMs in specialized domains such as mental health.

After fine-tuning, the models were quantized using QLoRA~\cite{qlora}, a process that enables efficient operation on consumer-grade hardware. This optimization was critical for deploying the fine-tuned models on Ollama, a framework designed for the lightweight yet high-performance execution of LLMs. Based on the predictions of the LLMs, OpenAI-gpt-oss LLM makes the final diagnosis of the mental illness. Custom prompts are used to instruct the OpenAI-gpt-oss Reasoning LLM to understand the context of the prediction. Based on the provided context and the predictions of the LLMs, the model makes the final diagnosis prediction. Platform performance is evaluated in three main areas: 1) Evaluation of LLM fine-tuning, 2) Diagnostic Performance of Fine-Tuned LLM Consortium, and 3) Diagnostic Reasoning Performance of the OpenAI-gpt-oss LLM.

\subsection{Evaluation of LLM Fine-Tuning}

This evaluation focuses on measuring the effectiveness of the fine-tuning process in improving the diagnostic accuracy of LLMs within the platform. Specifically, we evaluated the performance of the fine-tuned Llama-3 model in its ability to identify psychiatric symptoms and produce accurate diagnoses based on conversational data between psychiatrists and patients~\cite{training-validation-loss}.

\begin{figure}[H]
\centering{}
\includegraphics[width=5.0in]{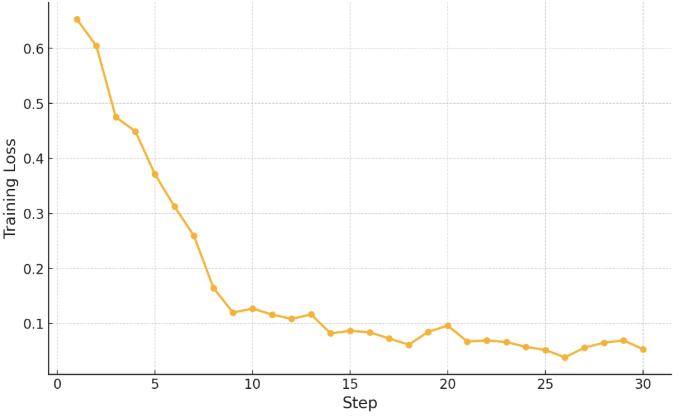}
\DeclareGraphicsExtensions.
\caption{Training loss during fine-tuning of the Llama-3 LLM}
\label{unsloth-tranning-loss}
\end{figure}
\begin{figure}[H]
\centering{}
\includegraphics[width=5.0in]{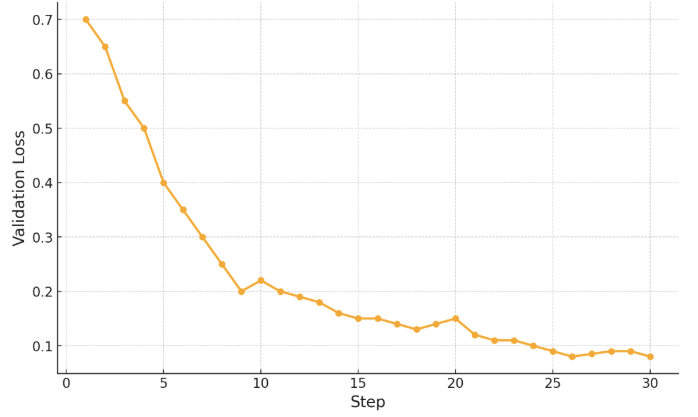}
\DeclareGraphicsExtensions.
\caption{Validation loss during fine-tuning of the Llama-3 LLM}
\label{unsloth-validation-loss}
\end{figure}
\begin{figure}[H]
\centering{}
\includegraphics[width=5.0in]{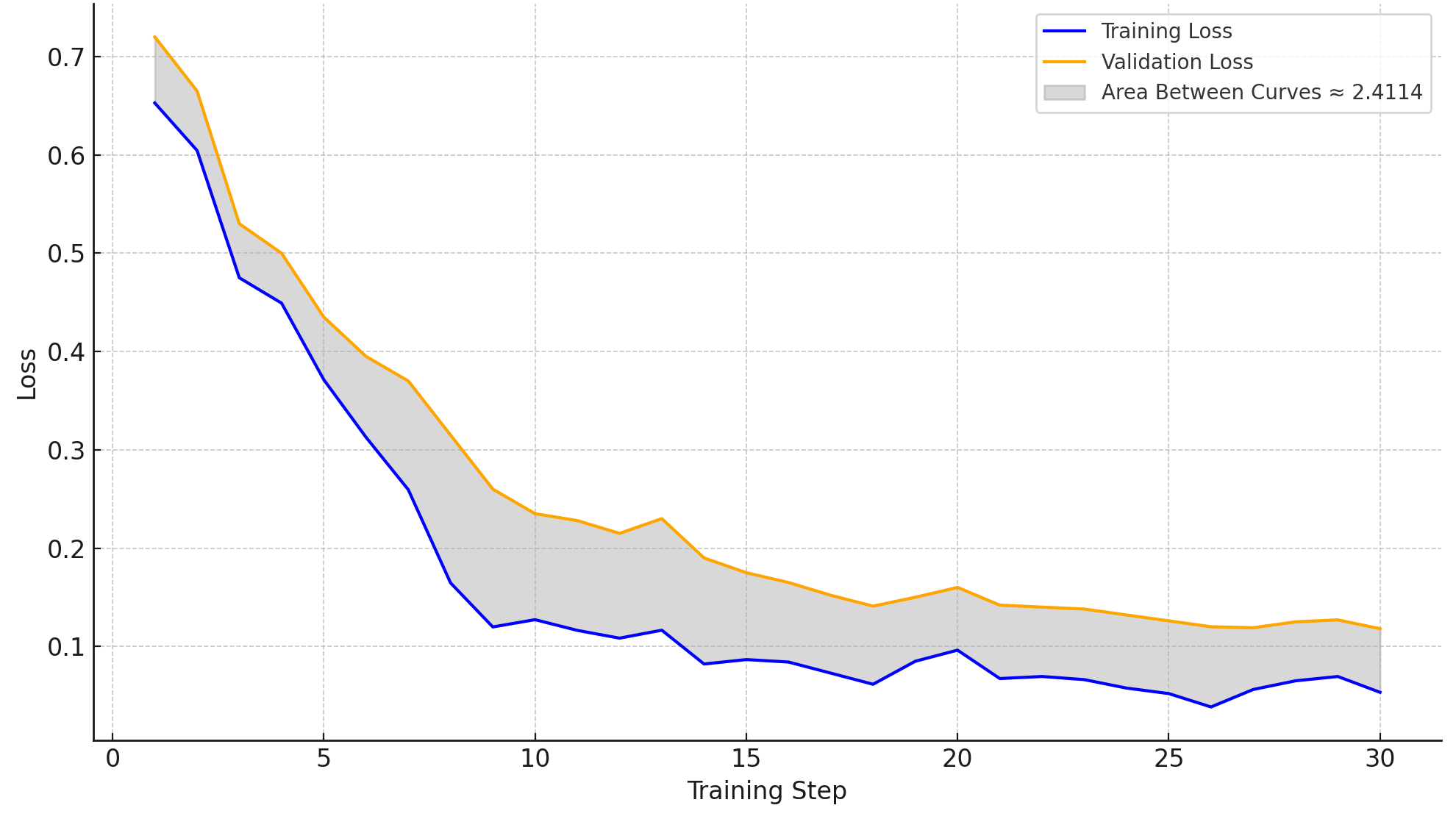}
\DeclareGraphicsExtensions.
\caption{Training vs. Validation Loss and Area Between Curves during Fine-Tuning of the Llama-3 LLM}
\label{unsloth-tranning-validation-loss}
\end{figure}

Throughout the fine-tuning process, we continuously monitored critical metrics—specifically, training loss and validation loss—to assess the model's learning dynamics and generalization ability~\cite{llm-finetune}. As shown in Figure~\ref{unsloth-tranning-validation-loss}, the training loss (Figure~\ref{unsloth-tranning-loss}) and validation loss (Figure~\ref{unsloth-validation-loss}) both exhibit a steep decline during the initial training steps, indicating rapid learning and effective assimilation of domain-specific patterns. The validation loss continues to decrease smoothly over time, stabilizing around step 25, which suggests improved generalization to unseen samples. Meanwhile, the training loss decreases more aggressively and plateaus slightly earlier, signaling convergence. Figure~\ref{unsloth-tranning-validation-loss} provides an integrated visualization of both metrics along with the area between the curves, which quantifies the generalization gap. This shaded area (\~2.41) highlights the difference between training and validation performance. The relatively narrow and consistently shrinking gap further confirms the model's ability to generalize well without overfitting. These trends collectively indicate that the fine-tuning process was effective and stable, enabling the LLM to adapt precisely to the psychiatric diagnostic domain while maintaining performance on unseen conversational data.

\begin{figure}[H]
\centering{}
\includegraphics[width=5.0in]{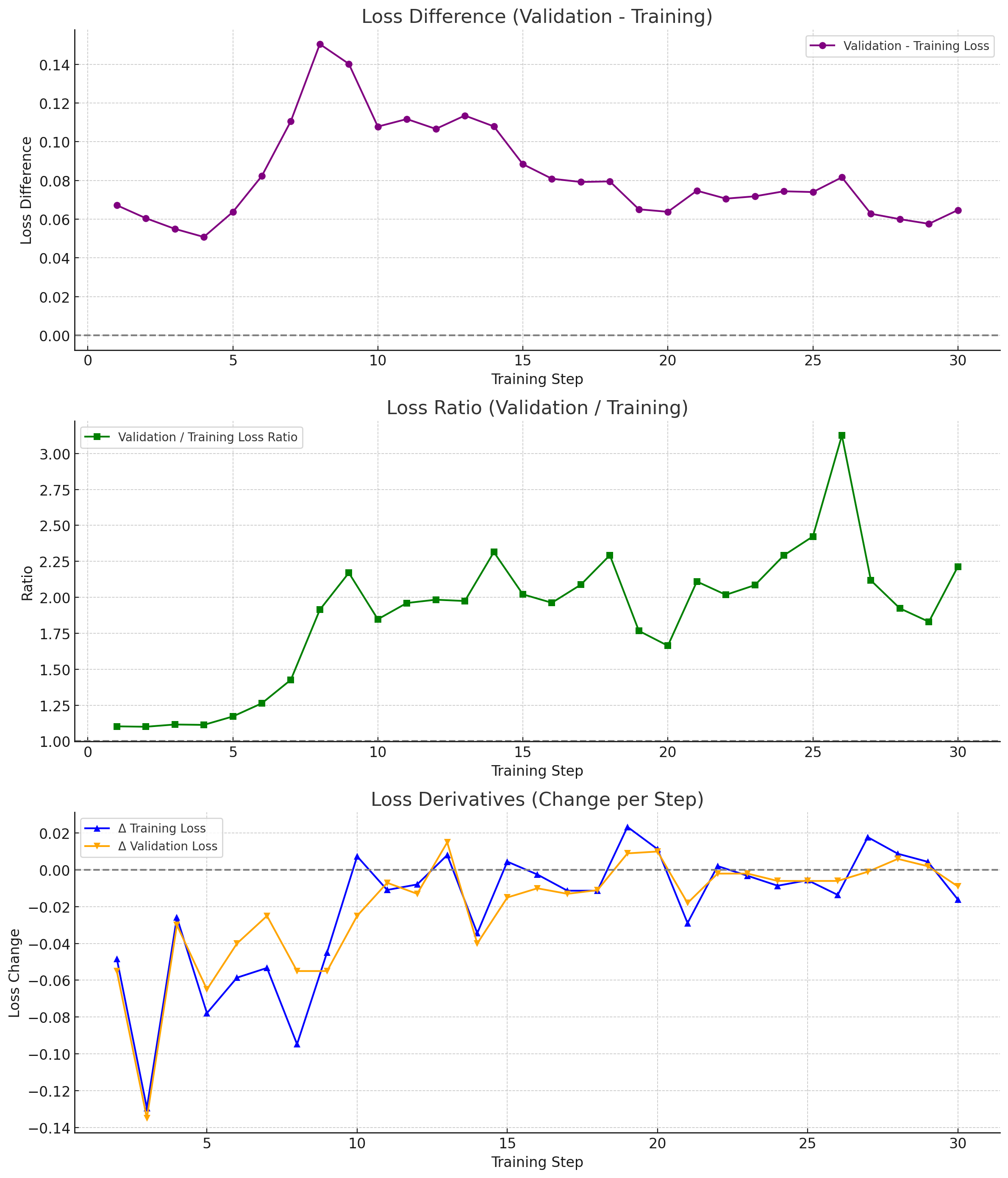}
\DeclareGraphicsExtensions.
\caption{Ratio of training to validation loss during the fine-tuning of the Llama-3 LLM.}
\label{unsloth-loss-ratio}
\end{figure}

Figure~\ref{unsloth-loss-ratio} captures multiple key training dynamics, including the loss difference, loss ratio, and loss derivatives over training steps, offering valuable insights into the model’s convergence behavior and generalization performance. The consistently positive loss difference (validation loss exceeding training loss) suggests signs of overfitting, especially at steps with noticeable spikes. The loss ratio, ranging from 1.0 to 3.0, highlights varying degrees of generalization, where a lower ratio reflects better alignment between training and validation performance. Additionally, the loss derivatives reveal rapid initial improvements followed by smaller, oscillating changes, indicating stabilization or saturation in the learning process~\cite{llm-loss-ratio}. 

\subsection{Diagnostic Performance of Fine-Tuned LLM Consortium}

Following the training phase, we assessed the predictive performance of the fine-tuned models in the context of psychiatric diagnosis. This evaluation involved comparing real diagnostic annotations—based on psychiatrist–patient conversations—with the predictions generated by both the baseline (pre-trained) LLMs and their fine-tuned counterparts. 

\begin{figure}[H]
\centering{}
\includegraphics[width=5.2in]{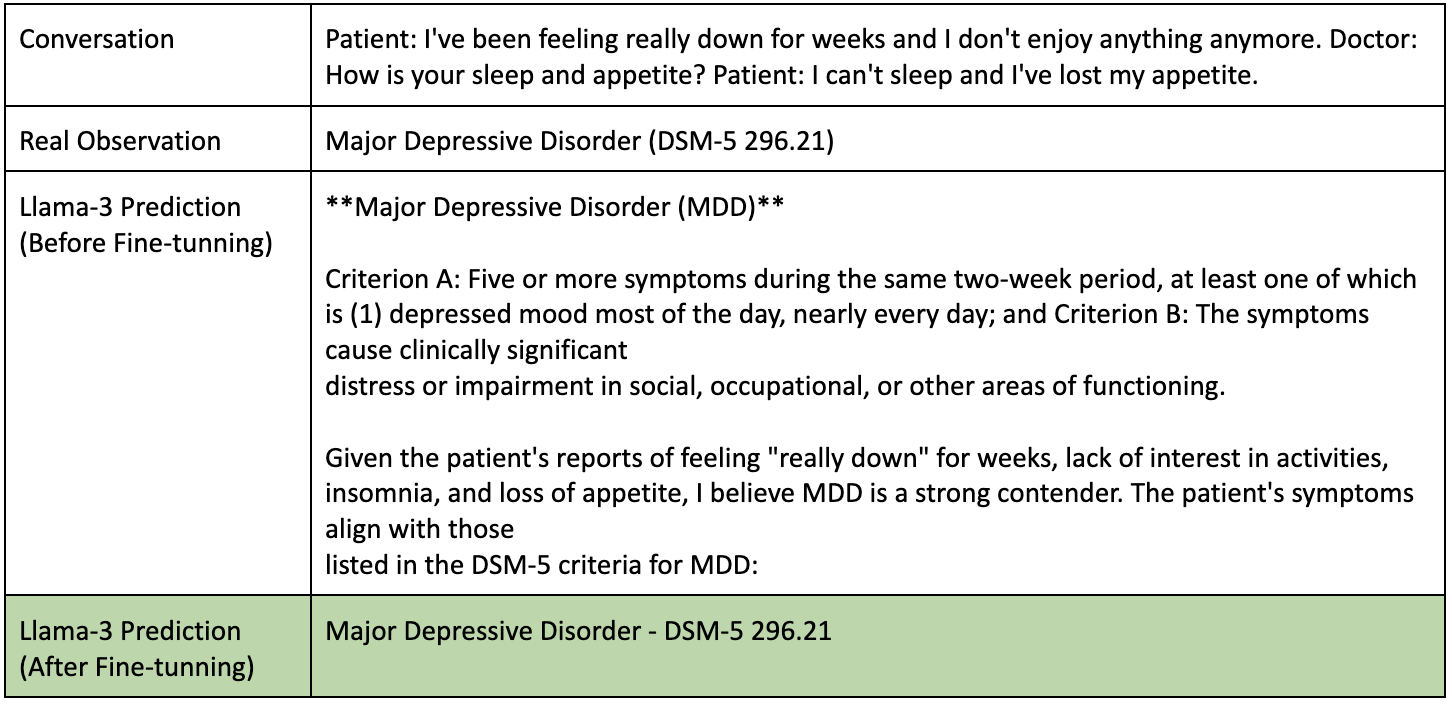}
\DeclareGraphicsExtensions.
\caption{Diagnostic output of the Llama-3 LLM for symptoms indicative of Major Depressive Disorder (DSM-5 296.21)}
\label{prediction-llama-v1}
\end{figure}

\begin{figure}[H]
\centering{}
\includegraphics[width=5.2in]{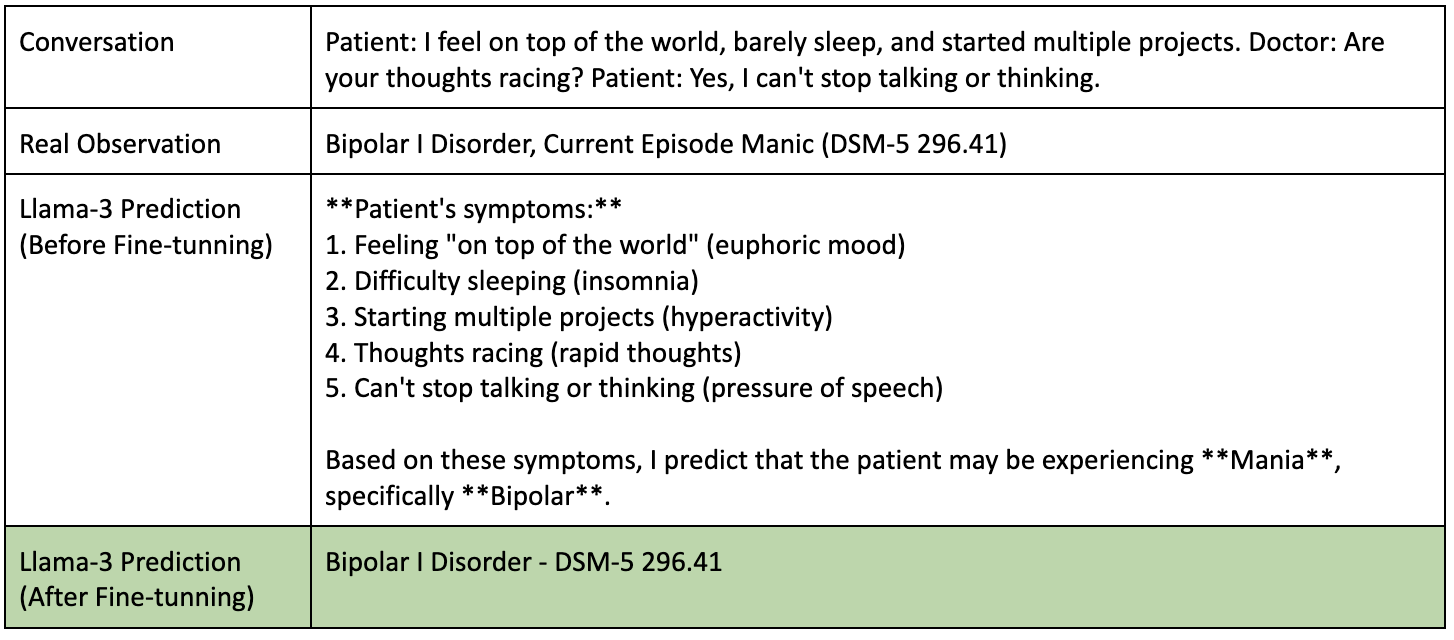}
\DeclareGraphicsExtensions.
\caption{Diagnostic output of the Llama-3 LLM for symptoms consistent with Bipolar I Disorder, Current Episode Manic (DSM-5 296.41)}
\label{prediction-llama-v2}
\end{figure}

Figure~\ref{prediction-llama-v1} and Figure~\ref{prediction-llama-v2} present the diagnostic predictions made by the Llama-3 model~\cite{llama-3, llama-4} before and after fine-tuning for two psychiatric conditions: Major Depressive Disorder (DSM-5 296.21) and Bipolar I Disorder, Current Episode Manic (DSM-5 296.41)~\cite{diagnosis-1}. Before fine-tuning, the model generated verbose but loosely structured outputs, relying on high-level symptom descriptions without explicitly mapping them to DSM-5 diagnostic codes~\cite{dsm-5}. While it identified relevant symptoms (e.g., euphoric mood, insomnia, and loss of appetite), it did not consistently align them with standardized diagnostic criteria. After fine-tuning on a domain-specific dataset, the model demonstrated significantly improved diagnostic precision. It accurately produced concise and clinically valid diagnoses with correct DSM-5 codes, directly inferred from the contextual patient-physician conversations. This highlights the effectiveness of targeted fine-tuning in enhancing clinical reasoning and diagnostic accuracy in mental health applications.

\begin{figure}[H]
\centering{}
\includegraphics[width=5.2in]{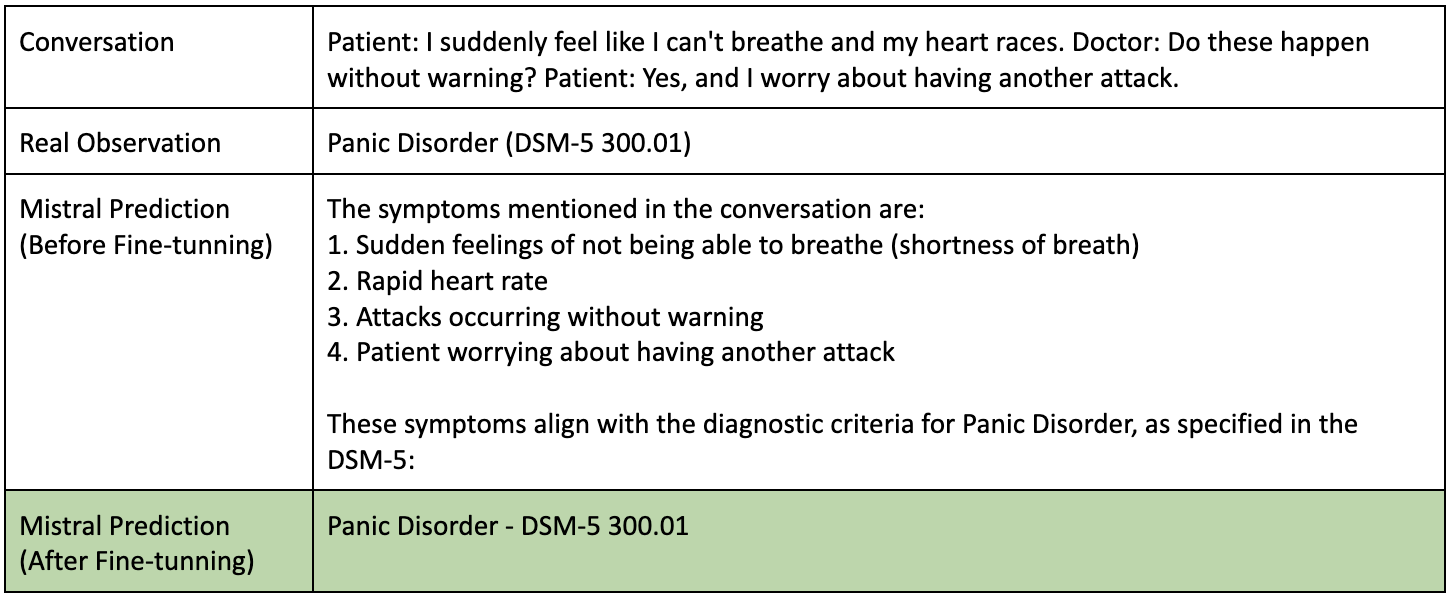}
\DeclareGraphicsExtensions.
\caption{Diagnostic output of the Mistral LLM for symptoms consistent with Panic Disorder (DSM-5 300.01)}
\label{prediction-mistral-v2}
\end{figure}

\begin{figure}[H]
\centering{}
\includegraphics[width=5.2in]{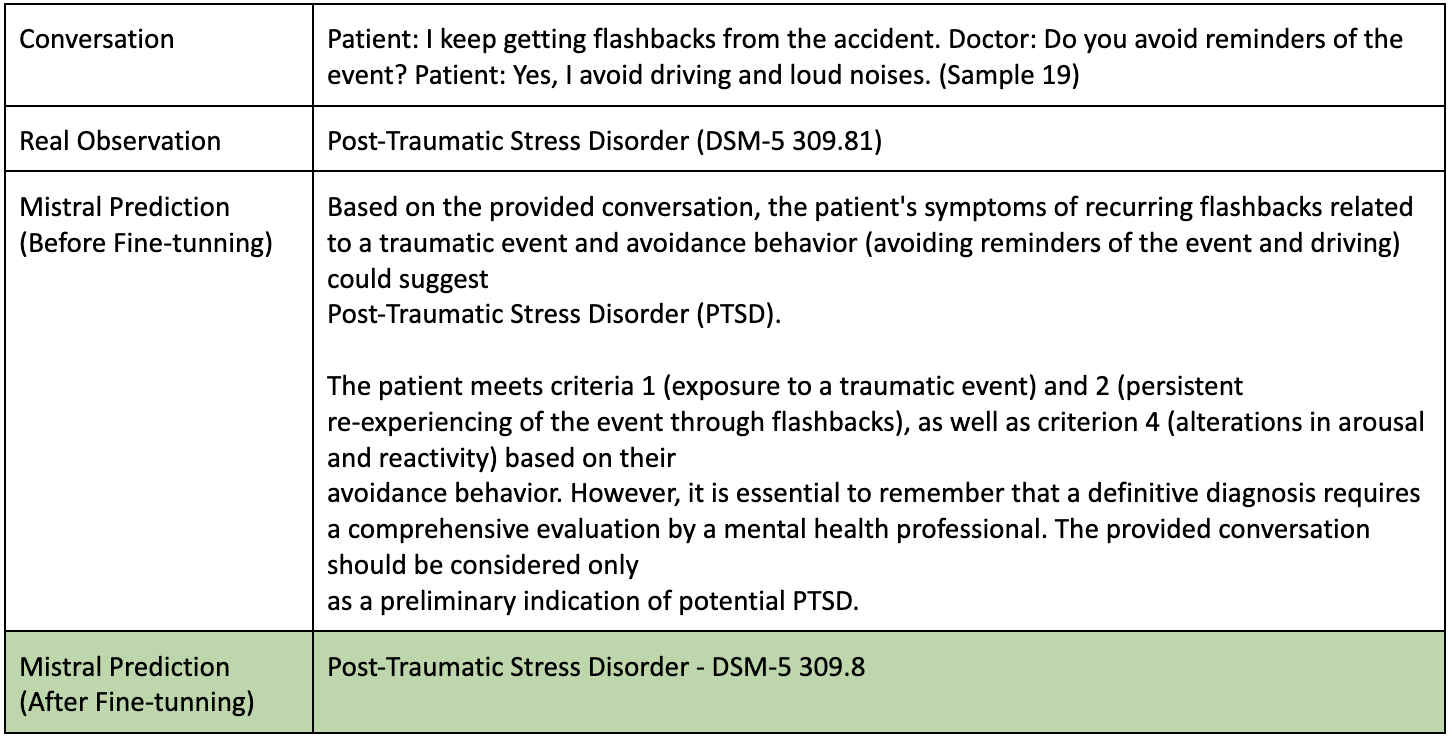}
\DeclareGraphicsExtensions.
\caption{Diagnostic output of the Mistral LLM for symptoms consistent with Post-Traumatic Stress Disorder (DSM-5 309.81)}
\label{prediction-mistral-v1}
\end{figure}

Figure~\ref{prediction-mistral-v1} and Figure~\ref{prediction-mistral-v2} illustrate the diagnostic predictions produced by the Mistral model~\cite{mistral-llm} for Panic Disorder and Post-Traumatic Stress Disorder (PTSD)~\cite{diagnosis-2}. Before fine-tuning, Mistral was able to recognize relevant clinical symptoms from the conversation and partially align them with DSM-5 criteria. For instance, in the panic disorder case, the model identified key features such as shortness of breath, rapid heart rate, and anticipatory anxiety about future attacks. Similarly, for PTSD, it recognized trauma exposure, intrusive flashbacks, and avoidance behaviors, citing appropriate DSM-5 criteria. However, after fine-tuning, the model demonstrated improved diagnostic specificity and accuracy. It produced concise and direct classifications that fully matched the DSM-5 diagnostic codes (e.g., DSM-5 300.01 for Panic Disorder and DSM-5 309.81 for PTSD)~\cite{dsm-5-criteria}. This confirms that fine-tuning enhanced Mistral’s capacity for structured clinical reasoning and reliable psychiatric diagnosis from patient-doctor dialogues.

\begin{figure}[H]
\centering{}
\includegraphics[width=5.2in]{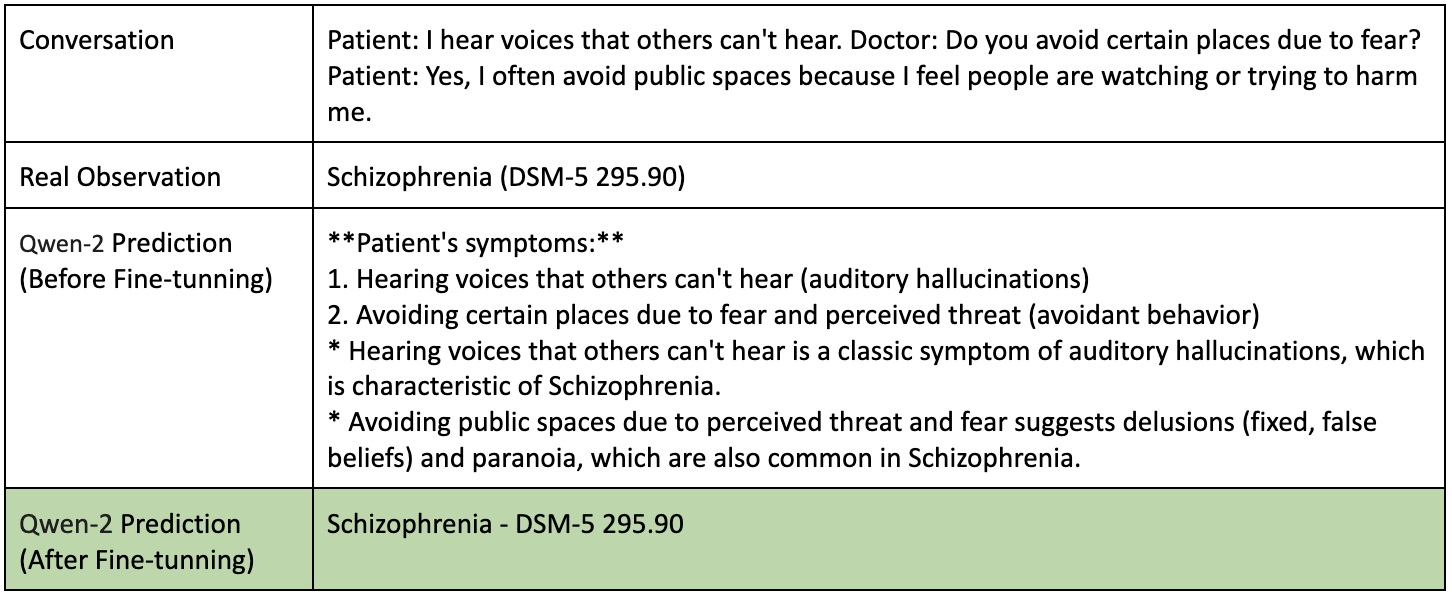}
\DeclareGraphicsExtensions.
\caption{Diagnostic output of the Qwen2 LLM for symptoms consistent with Schizophrenia (DSM-5 295.90)}
\label{prediction-qwen-v2}
\end{figure}

\begin{figure}[H]
\centering{}
\includegraphics[width=5.2in]{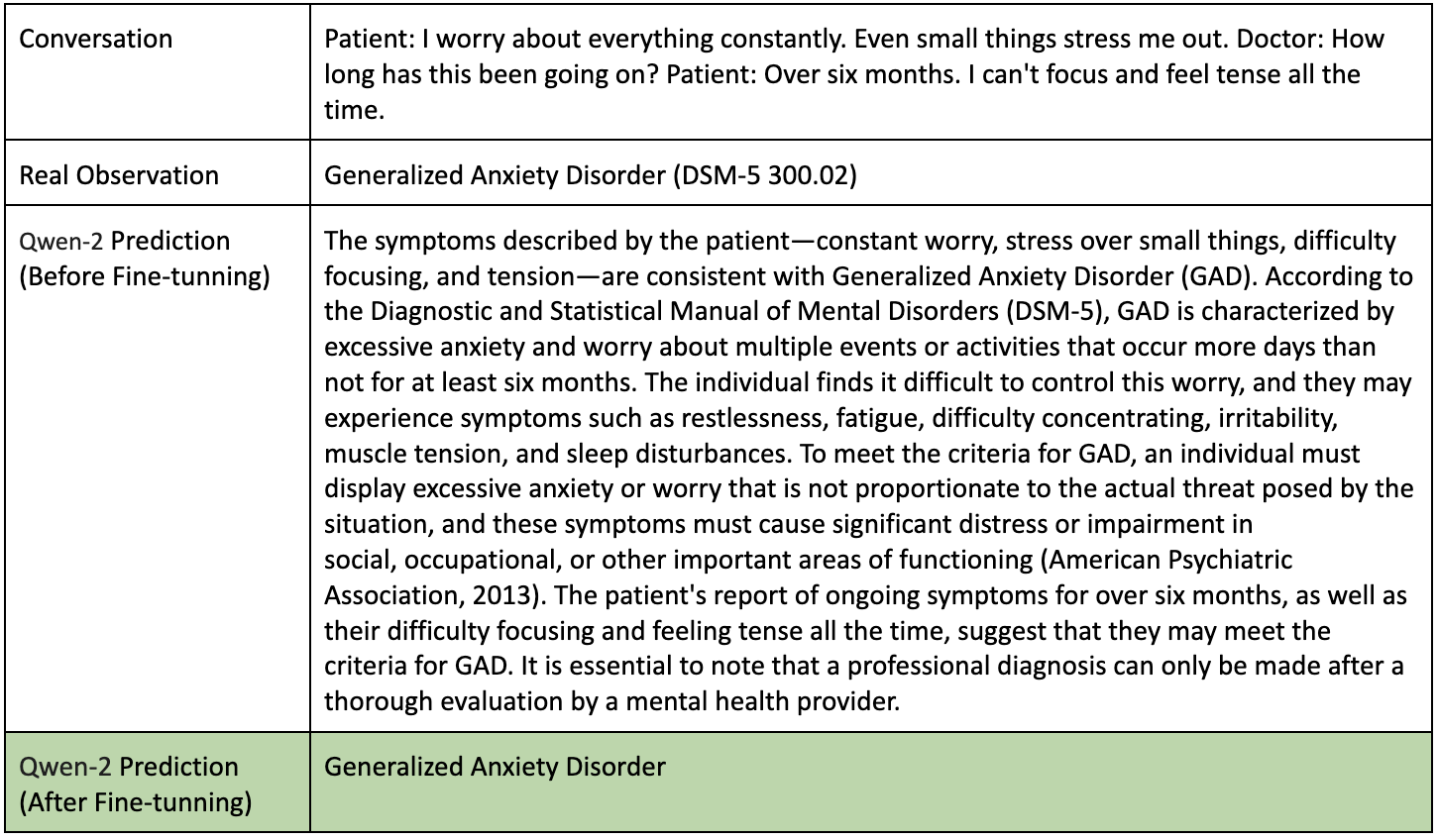}
\DeclareGraphicsExtensions.
\caption{Diagnostic output of the Qwen2 LLM for symptoms indicative of Generalized Anxiety Disorder (DSM-5 300.02)}
\label{prediction-qwen-v1}
\end{figure}

Figure~\ref{prediction-qwen-v1} and Figure~\ref{prediction-qwen-v2} present the diagnostic outputs generated by the Qwen2 model~\cite{on-device-qwen2} for Schizophrenia and Generalized Anxiety Disorder, respectively~\cite{diagnosis-3}. Prior to fine-tuning, the model identified relevant symptoms and attempted to match them to DSM-5 criteria. However, the predictions were often verbose, interpretative, and occasionally lacked clinical precision. After fine-tuning on the psychiatric dataset, the Qwen2 model accurately mapped symptom clusters to their corresponding DSM-5 diagnoses with greater clarity and conciseness, yielding outputs that are more aligned with psychiatric clinical standards.

These results demonstrate that the fine-tuned models consistently produce predictions that closely align with clinically validated diagnoses, showing improved precision, consistency, and interpretability. Compared to their baseline versions, the fine-tuned LLMs exhibit a substantial increase in diagnostic accuracy, underscoring the effectiveness of task-specific fine-tuning in mental health applications. These findings validate the utility of LLMs as decision-support tools in AI-assisted psychiatric diagnosis.

\subsection{Diagnostic Reasoning Performance of the OpenAI-gpt-oss LLM}

\begin{figure}[H]
\centering{}
\includegraphics[width=5.2in]{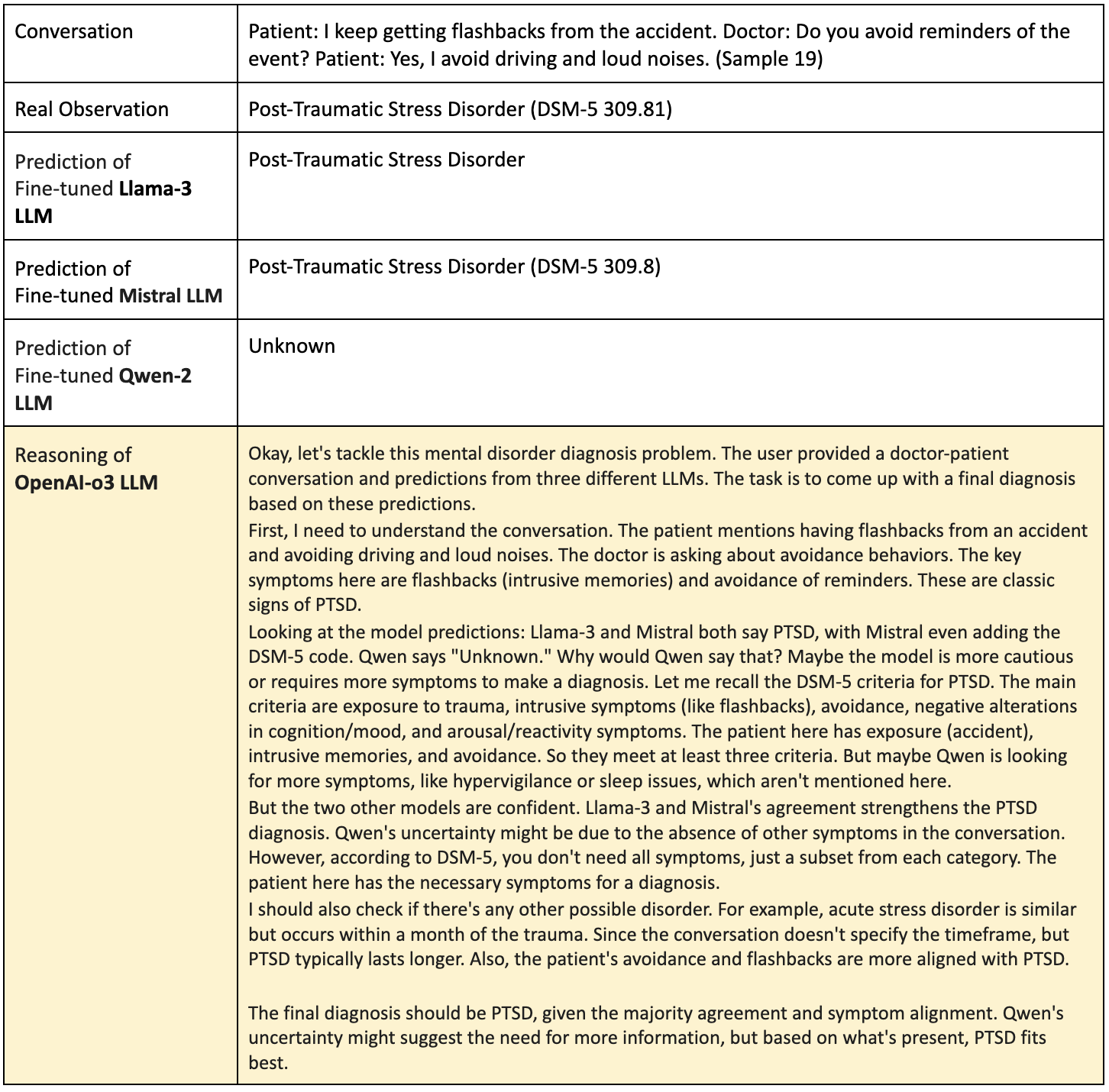}
\DeclareGraphicsExtensions.
\caption{Final diagnostic reasoning produced by the OpenAI-gpt-oss LLM through multi-model consensus.}
\label{prediction-o3}
\end{figure}

In this evaluation, we assessed the diagnostic reasoning performance of the OpenAI-gpt-oss reasoning LLM by comparing its final diagnosis predictions with those generated independently by multiple fine-tuned LLMs. The goal was to evaluate OpenAI-gpt-oss’s ability to synthesize diverse diagnostic outputs and determine the most accurate and clinically appropriate outcome. Figure~\ref{prediction-o3} illustrates a comparative analysis of the diagnostic predictions from the Llama-3, Mistral, and Qwen2 models, alongside the final output produced by OpenAI-gpt-oss. The figure highlights the reasoning model’s ability to interpret and reconcile varying predictions, applying structured clinical logic to arrive at a DSM-5-aligned diagnosis. The results demonstrate that OpenAI-gpt-oss significantly enhances diagnostic reliability by evaluating the coherence and clinical relevance of the LLM consortium’s outputs. This consensus-driven reasoning step adds an important layer of interpretability and robustness, reinforcing the value of integrating a dedicated reasoning LLM within the framework for AI-assisted psychiatric diagnosis.

\section{Related Work}

Recent advances have seen LLMs and multimodal AI systems increasingly applied across a wide range of healthcare domains. These efforts collectively highlight the expanding capabilities of LLMs in clinical reasoning, decision support, and medical knowledge retrieval. Despite these advancements, most prior work remains focused on general medicine or physical health domains, with limited specialization in psychiatric evaluation or adherence to standardized diagnostic frameworks such as the DSM‑5. Furthermore, while individual models often perform well in isolated tasks, they rarely incorporate structured reasoning or multi-model coordination—critical components for nuanced psychiatric assessments. Table~\ref{t_bc_platforms} presents a comparative analysis of these prior systems in relation to our proposed framework. In the subsections that follow, we provide a detailed overview of each relevant system’s architecture, capabilities, and relevance to psychiatric diagnostic support.


\subsection{Med‑PaLM}
The Med‑PaLM family of models, including Med‑PaLM and Med‑PaLM-2, developed by Google Research, represents some of the most advanced and widely evaluated LLMs for clinical and biomedical tasks~\cite{med-palm}. These models are fine-tuned versions of the PaLM (Pathways Language Model) foundation model, tailored to the medical domain through instruction tuning using a combination of publicly available medical datasets, expert-annotated clinical examples, and proprietary benchmarks. 

Med‑PaLM-2 demonstrated expert-level competency, achieving over 85\% accuracy on U.S. Medical Licensing Examination (USMLE)–style multiple-choice questions, including datasets such as MedQA, MedMCQA, and PubMedQA. Notably, in head-to-head evaluations, physicians preferred Med‑PaLM-2’s long-form answers to those written by human experts across multiple dimensions, including factual correctness, comprehensiveness, and safety. The model also performed well on several open-ended generation tasks, such as consumer health question answering and summarization of clinical information.

In addition to the text-only versions, a multimodal extension called Med‑PaLM-M was developed, enabling the model to jointly reason over textual and visual inputs. This includes radiology images, dermatology photos, pathology slides, and ophthalmic scans. Evaluated on the MultiMedBench benchmark, Med‑PaLM-M demonstrated state-of-the-art performance across 14 diverse tasks encompassing classification, retrieval, and open-ended question answering with multimodal inputs. The model is capable of interpreting complex visual patterns and aligning them with clinical narratives, allowing for integrated diagnostic reasoning that mirrors specialist workflows.

Furthermore, the Med‑PaLM series introduces an ethical framework and rigorous safety assessments for AI in healthcare. This includes human evaluation protocols to assess bias, hallucination risks, and alignment with clinical guidelines. Despite its strengths, limitations remain—such as interpretability challenges and dependence on high-quality supervised data—underscoring the need for hybrid frameworks that can further ensure trust and safety in clinical applications.

\subsection{LLM for Differential Diagnosis (DDx)}
The ``LLM for DDx'' framework~\cite{llm-ddx} investigates the application of LLMs to the task of generating differential diagnoses from clinical vignettes, including both structured patient data and free-text clinical notes. By leveraging prompt engineering techniques, the system guides the LLM to output ranked diagnostic hypotheses based on symptomatology, patient history, and clinical context. This framework adopts few-shot or zero-shot prompting, often augmented with chain-of-thought reasoning or self-consistency sampling, to simulate a clinician’s diagnostic reasoning. The model is evaluated on publicly available datasets such as MedQA, MIMIC-III case notes, and Clinical Case Challenge benchmarks. It demonstrates promising capabilities in recognizing comorbidities, parsing temporal disease progression, and identifying rare or underrepresented conditions—tasks that often challenge traditional rule-based decision support systems. One of the key strengths of this approach lies in its adaptability: the model can generalize across specialties (e.g., internal medicine, neurology, pediatrics) without needing task-specific retraining. However, the system does not integrate multimodal data sources (e.g., lab results, imaging), and lacks structured inference mechanisms like probabilistic reasoning or iterative refinement via specialized LLMs. As such, its diagnostic accuracy can be sensitive to prompt phrasing and input variability, necessitating human-in-the-loop oversight for safe clinical deployment. Despite these limitations, the work highlights the potential of LLMs to augment differential diagnosis in resource-constrained or high-ambiguity settings, offering a foundation for future systems that incorporate more structured reasoning pipelines and multimodal data fusion.

\subsection{Me‑Llama}
Me‑Llama~\cite{me-llama} adapts the Llama architecture for biomedical applications through a two-phase process: (1) continual pretraining on large-scale biomedical corpora—including PubMed articles, clinical guidelines, and de-identified electronic health records (EHRs); and (2) instruction tuning using curated clinical dialogues and task-specific prompts aligned with workflows typical in patient care.  The model is optimized for downstream tasks such as clinical note summarization, medical question answering, and evidence extraction from unstructured clinical documents. Benchmarks show that Me‑Llama outperforms general-purpose LLMs on several biomedical NLP tasks, including the BioASQ challenge, PubMedQA, and MedNLI. Despite being limited to text-only inputs (i.e., it does not incorporate imaging, waveform, or wearable sensor data), Me‑Llama exhibits a strong semantic understanding of biomedical terminology, abbreviations, and guideline-referenced medical reasoning. Its lightweight fine-tuning also makes it suitable for deployment in edge healthcare systems, such as point-of-care mobile applications and hospital EMR-integrated tools. Overall, Me‑Llama demonstrates the feasibility and performance benefits of domain-specific LLM adaptation within clinical language environments.

\subsection{DrHouse}
DrHouse~\cite{drhouse} introduces an advanced virtual provider assistant system that integrates LLM-based diagnostic reasoning with real-time physiological data collected from consumer-grade wearable sensors (e.g., smart watches, sleep trackers, fitness monitors). The system continuously monitors health signals such as heart rate variability, blood oxygen saturation (SpO\textsubscript{2}), sleep quality, step count, and circadian rhythm alignment, enabling longitudinal health tracking and context-aware decision-making. In contrast to static, prompt-only systems, DrHouse engages users through multi-turn, adaptive dialogues that emulate the back-and-forth of a clinical consultation. The LLM component dynamically adjusts its diagnostic hypotheses using an iterative concurrent reasoning framework—recalculating disease likelihoods as new data or clarifications are introduced. Additionally, DrHouse retrieves up-to-date medical knowledge from external expert databases such as UpToDate, PubMed abstracts, and clinical guidelines, thereby ensuring that recommendations remain grounded in the latest medical evidence.

The system employs a dual-loop architecture that balances real-time inference with long-term patient modeling. It maintains an evolving profile of each user’s medical baseline and incorporates Bayesian-style updates for probabilistic diagnosis refinement. DrHouse was evaluated on both public benchmark datasets (e.g., MedQA, Symptoma) and proprietary longitudinal datasets derived from wearable telemetry and self-reported health surveys. Quantitatively, the model achieved up to an 18.8\% improvement in diagnostic accuracy compared to baseline LLMs without sensor fusion. Qualitatively, user studies showed that 91.7\% of patients found the interaction intuitive and trustworthy, while 75\% of clinicians expressed confidence in its ability to support primary care triage. These results underscore the promise of multimodal, sensor-aware agentic systems for scalable, AI-assisted frontline healthcare.

\subsection{CDSS}
CDSS~\cite{CDSS} introduces a novel Clinical Decision Support System tailored for mental health diagnostics by synergistically combining LLMs with constraint logic programming (CLP). The system ingests natural language diagnostic manuals—specifically DSM‑5‑TR and ICD‑11 CDDR—and uses an LLM to transform each diagnostic criterion into logic rules (e.g., Datalog clauses)~\cite{CDSS}. These candidate rules are then vetted and refined by domain experts to ensure clinical fidelity before being executed by an off-the-shelf CLP engine to derive patient-specific diagnoses based on structured patient data. 

In empirical evaluations, the hybrid CDSS is compared against two baselines: an “LLM-only” approach that directly generates diagnostic outputs, and an intermediate LLM-to-CLP pipeline without expert oversight. Results indicate that only the expert-validated pipeline consistently produces diagnoses aligned with official manuals, highlighting the necessity of human-in-the-loop rule verification to prevent hallucinations and maintain interpretability. The authors also emphasize operational benefits: the logic rules are transparent and inspectable, facilitating clinician trust and auditability. Moreover, the approach addresses critical ethical concerns by avoiding the direct ingestion of sensitive patient data into the LLM—patient records are instead processed via the CLP engine, mitigating privacy and safety risks associated with raw LLM consumption. 

This work represents a significant advancement in mental health AI, as it operationalizes a structured, interpretable, and demonstrably safe CDSS anchored in expert-validated logic and modular LLM capabilities—marking a concrete step toward real-world psychiatric diagnostic tools.

\subsection{Weda‑GPT}
Weda‑GPT~\cite{wedagpt} is a culturally-informed clinical decision-support system that leverages fine-tuned Llama‑3 models to provide diagnostic assistance and therapeutic recommendations within indigenous and traditional medicine frameworks. Designed specifically for use in the Indonesian archipelago, Weda‑GPT incorporates linguistic, cultural, and epistemological knowledge derived from indigenous medical texts, oral traditions, and local practitioner expertise.

The system is built using a multi-stage fine-tuning pipeline: starting from a Llama‑3 base model, it is further adapted with region-specific datasets encompassing herbal pharmacology, traditional syndromic classifications, and culturally embedded health beliefs. Special emphasis is placed on aligning model outputs with culturally appropriate terminology and explanatory models, enabling the system to provide contextually sensitive health advice.

Weda‑GPT has been evaluated through case-based testing and participatory design sessions involving local healers and community health workers. The results show that the model effectively maps patient symptoms to culturally relevant diagnoses and treatments, including herbal prescriptions and ritual-based healing practices. Moreover, users reported a high degree of trust and interpretability in the system’s responses, in part due to its capacity to explain recommendations in locally meaningful terms.

Although Weda‑GPT does not operate within Western psychiatric diagnostic categories such as DSM‑5, it highlights the broader potential of LLMs to support non-Western health systems and pluralistic medical epistemologies. Its design underscores the importance of cultural adaptation and domain-specific alignment when deploying AI in diverse global health contexts. As such, Weda‑GPT serves as a complementary model to Western-centric clinical decision systems, demonstrating the scalability and flexibility of LLM-based healthcare tools across sociocultural boundaries.

\begin{table*}[!htb]\centering
\vspace{0.1in}
\caption {LLM-based Medical Diagnosis Framework Comparison}
\begin{adjustbox}{width=1\textwidth}
\label{t_bc_platforms}
\begin{tabular}{lccccccc}
\toprule
\thead{Platform} & \thead{Domain} & \thead{Fine-tuning\\Support} & \thead{Running LLM} & \thead{Vision LM\\Support} & \thead{Reasoning LLM\\Support} & \thead{LLM Consortium\\Support} \\
\midrule
Psychiatric-Diagnoses & Psychiatric & \cmark & Llama-3, Mistral, Qwen-2 & \xmark & \cmark & \cmark \\
Med-PaLM~\cite{med-palm} & General medicine & \cmark & PaLM & \xmark & \xmark & \xmark \\
LLM for DDx~\cite{llm-ddx} & General medicine & \cmark & Not specified & \xmark & \cmark & \xmark \\
Me-LLaMA~\cite{me-llama} & General medicine & \cmark & Llama & \xmark & \xmark & \xmark \\
CDSS~\cite{CDSS} & Mental Health & \xmark & GPT-4 & \xmark & \xmark & \xmark \\
DrHouse~\cite{drhouse} & General medicine & \cmark & Not specified & \cmark & \cmark & \xmark \\
Weda-GPT~\cite{wedagpt} & Indigenous Medicine & \cmark & Llama-3 & \xmark & \xmark & \xmark \\
\bottomrule
\end{tabular}
\end{adjustbox}
\end{table*}

\section{Conclusions and Future Work}

In this paper, we present an AI-assisted diagnostic framework that integrates a consortium of fine-tuned LLMs with a reasoning LLM (OpenAI-gpt-oss) to improve the accuracy, consistency, and transparency of psychiatric diagnosis while upholding Responsible AI principles. Recognizing the inherent subjectivity and variability in traditional mental health assessments—often based on unstructured clinical interviews—we proposed a novel architecture that leverages conversational data, custom prompt engineering, and multi-model consensus reasoning to replicate and improve upon clinical diagnostic workflows. The platform is structured into four key layers: the Data Lake Layer for managing annotated psychiatrist–patient dialogues; the fine-tuned LLM Layer for training models specialized in symptom analysis; the LLM Agent Layer for orchestrating model interactions and prompt generation; and the OpenAI-gpt-oss Reasoning Layer, which synthesizes model outputs into a final, reliable diagnosis aligned with DSM-5 criteria. Our approach demonstrates that AI systems, when trained and orchestrated properly, can support mental health professionals by offering data-driven insights and reducing diagnostic variability. The use of low-rank adapters and quantization techniques further enables efficient deployment on consumer-grade hardware, making the system accessible in real-world clinical and remote care settings. To the best of our knowledge, this research represents the first end-to-end integration of fine-tuned large language models (LLMs) with a reasoning engine to standardize psychiatric diagnoses. It lays the foundation for future advancements in AI-assisted eHealth systems, where intelligent agents can augment clinical decision-making while preserving interpretability and ethical responsibility. Future work will focus on clinical validation, multilingual adaptation, and integration with multimodal inputs—such as voice, facial expressions, and affective signals—to enhance diagnostic depth, contextual understanding, and empathy.


\section*{Acknowledgements}

This work was supported in part by the DoD Center of Excellence in AI and Machine Learning (CoE-AIML) under Contract Number W911NF-20-2-0277 with the U.S. Army Research Laboratory. 

\bibliographystyle{IEEEtran}
\bibliography{reference}

\pagebreak

\section*{Authors}

\begin{wrapfigure}{L}{0.2\textwidth}
\centering
\includegraphics[width=0.2\textwidth]{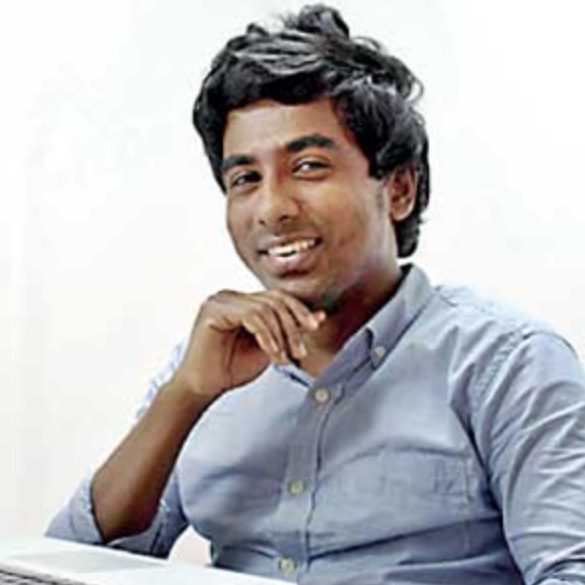}
\end{wrapfigure}


\textbf{Eranga Bandara} is a Senior Research Scientist at the Virginia Modeling, Analysis, and Simulation Center (VMASC), Old Dominion University, Virginia, USA. He done his Ph.D. in Computer Science with a specialization in blockchain and distributed systems. Following his doctoral work, he developed a deep interest in neuroscience and is currently pursuing academic studies in the Fundamentals of Neuroscience program at Harvard University. His interdisciplinary research interests span privacy-preserving AI, distributed systems, blockchain architectures, and next-generation wireless networks (5G/6G), with a growing focus on brain modeling, the application of artificial intelligence in neuroscience, and leveraging neuroscience principles to advance artificial intelligence.
\\
\\

\begin{wrapfigure}{L}{0.2\textwidth}
\centering
\includegraphics[width=0.2\textwidth]{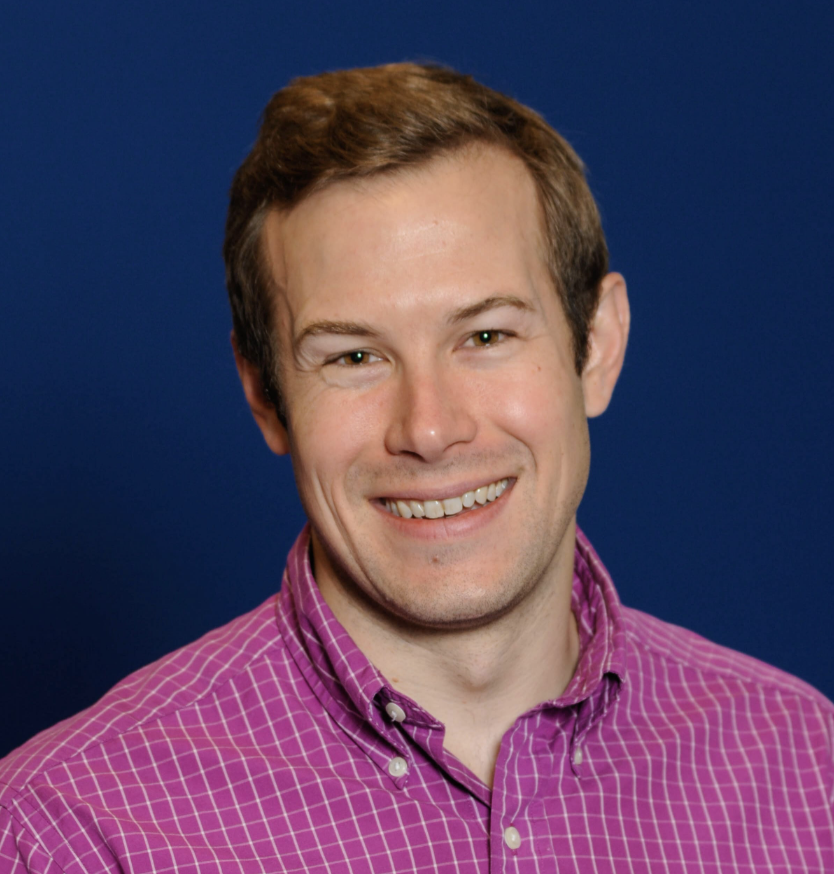}
\end{wrapfigure}

\textbf{Dr. Ross Gore} is a Research Associate Professor at the Virginia Modeling, Analysis and Simulation Center (VMASC) at Old Dominion University, Suffolk, Virginia, USA. He earned his Ph.D. (2012) and M.S. (2007) in Computer Science from the University of Virginia and his B.S. (2003) in Computer Science from the University of Richmond.  Dr. Gore’s research focuses on data science, predictive analytics, and simulation validation, with applications spanning public health, city planning, cybersecurity, and risk assessment. He is particularly interested in leveraging diverse data sources—from mobile devices to social media—to inform critical decision-making. Recent projects include using mobile phone data to inform public health policies during the COVID‑19 pandemic, applying social media analytics to explain geographic variations in obesity, and customizing cyber vulnerability assessments—often developed through close collaboration with stakeholders and iterative prototyping.  
\\
\\

\newpage

\begin{wrapfigure}{L}{0.2\textwidth}
\centering
\includegraphics[width=0.2\textwidth]{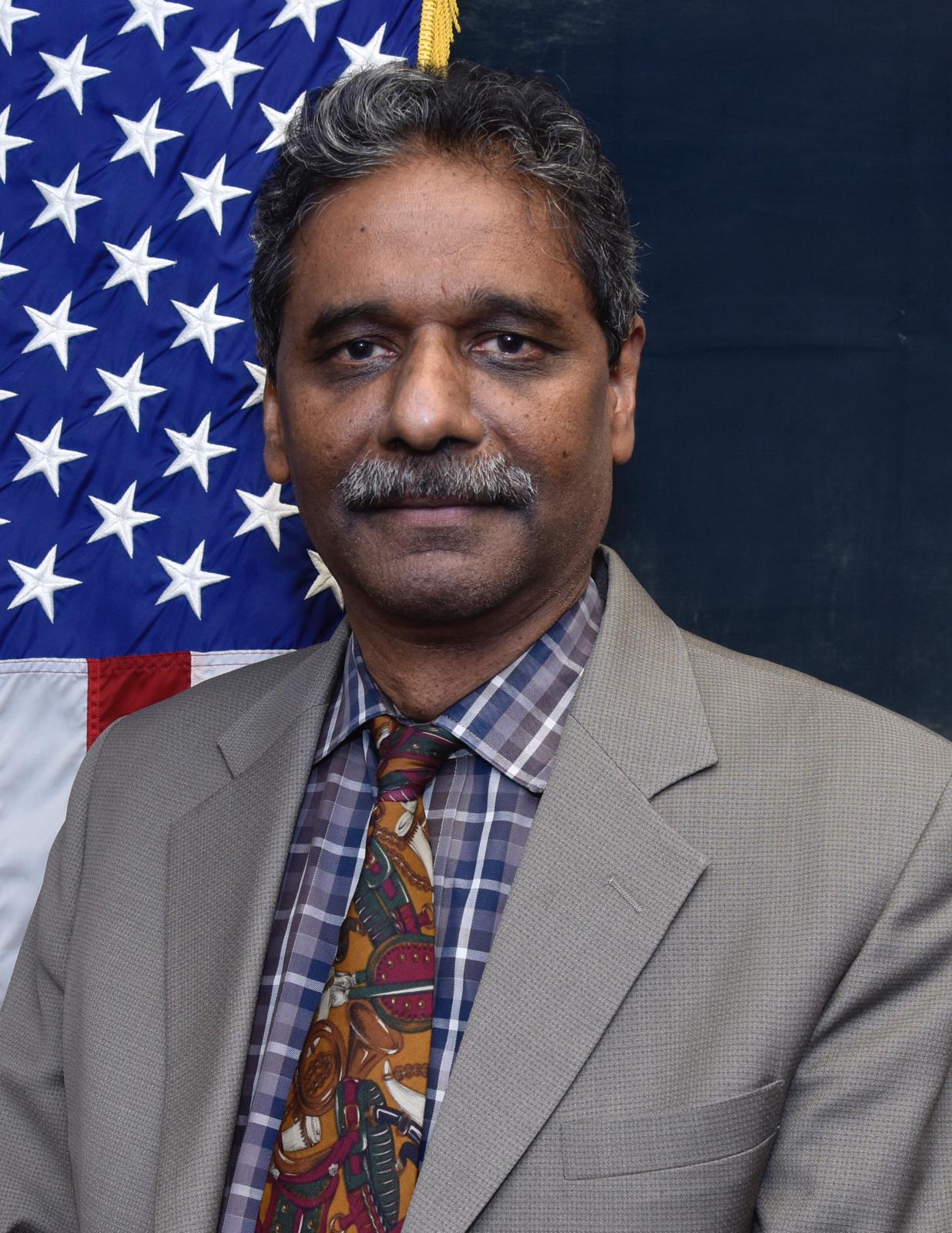}
\end{wrapfigure}

\textbf{Dr.\ Atma Ram Yarlagadda, MD} is an experienced psychiatrist and Installation Director of Psychological Health at McDonald Army Health Center, Fort Eustis, Virginia. He earned his medical degree from Danylo Halytsky Lviv State Medical University in 1983 and has over 40 years of clinical practice in military psychiatry. His work focuses on advancing mental health services for military personnel, emphasizing evidence-based assessment and treatment of mood and trauma-related disorders. 
\\
\\

\begin{wrapfigure}{L}{0.2\textwidth}
\centering
\includegraphics[width=0.2\textwidth]{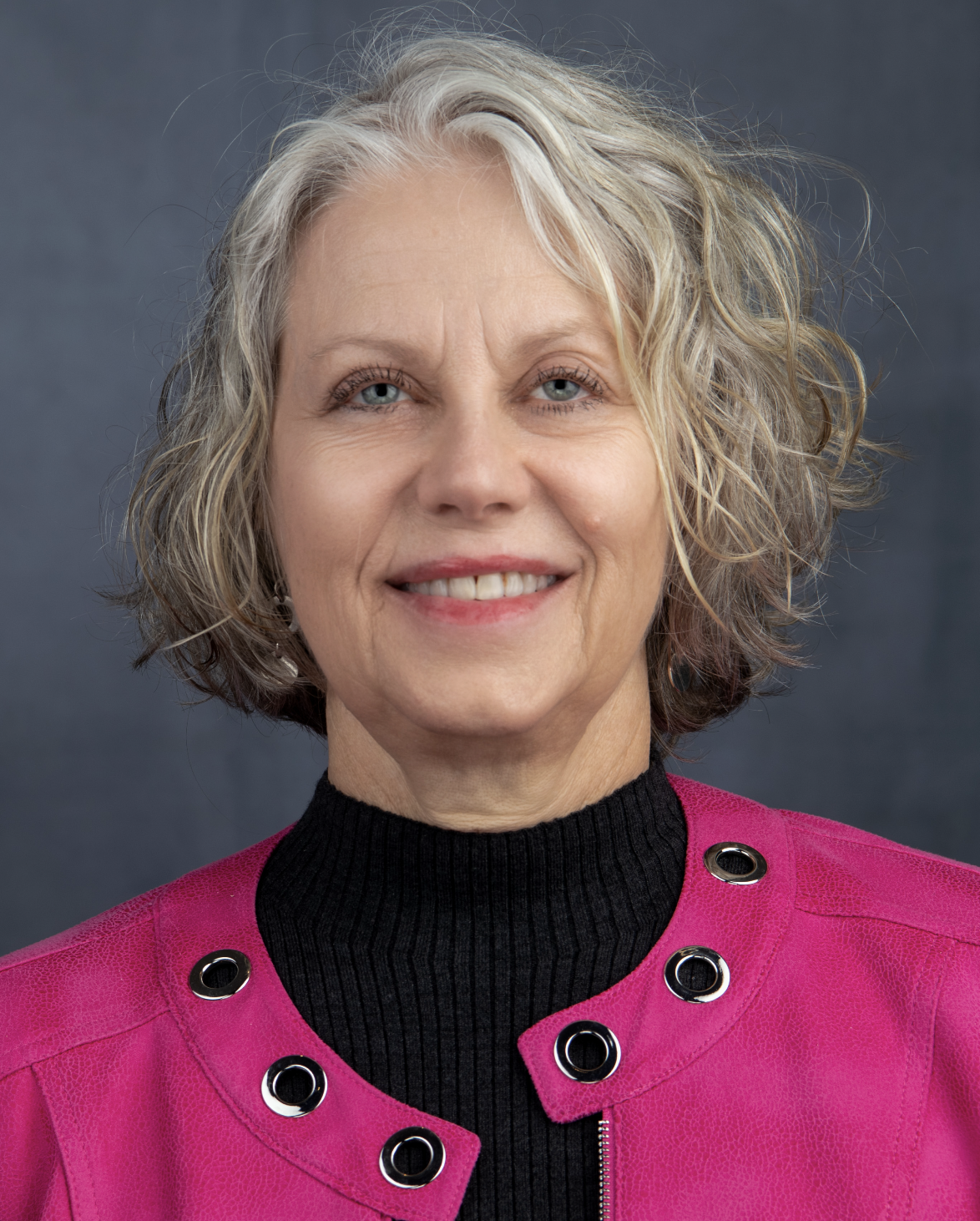}
\end{wrapfigure}

\textbf{Dr.\ Anita H.\ Clayton, MD} is the Wilford W. Spradlin Professor and Chair of the Department of Psychiatry and Neurobehavioral Sciences at the University of Virginia School of Medicine. She completed her MD and psychiatry residency at UVA, followed by service in the U.S. Navy Medical Corps before joining UVA faculty in 1990. Dr. Clayton is a leading expert in major depressive disorder, women’s mental health, sexual dysfunction, and reproductive psychiatry with over 225 peer-reviewed publications. She pioneered validated assessment tools such as the CSFQ, SIDI-F and DSDS and served as president of the International Society for the Study of Women’s Sexual Health. She is the current president of the American Society of Clinical Psychopharmacology. 
\\
\\


\begin{wrapfigure}{L}{0.2\textwidth}
\centering
\includegraphics[width=0.2\textwidth]{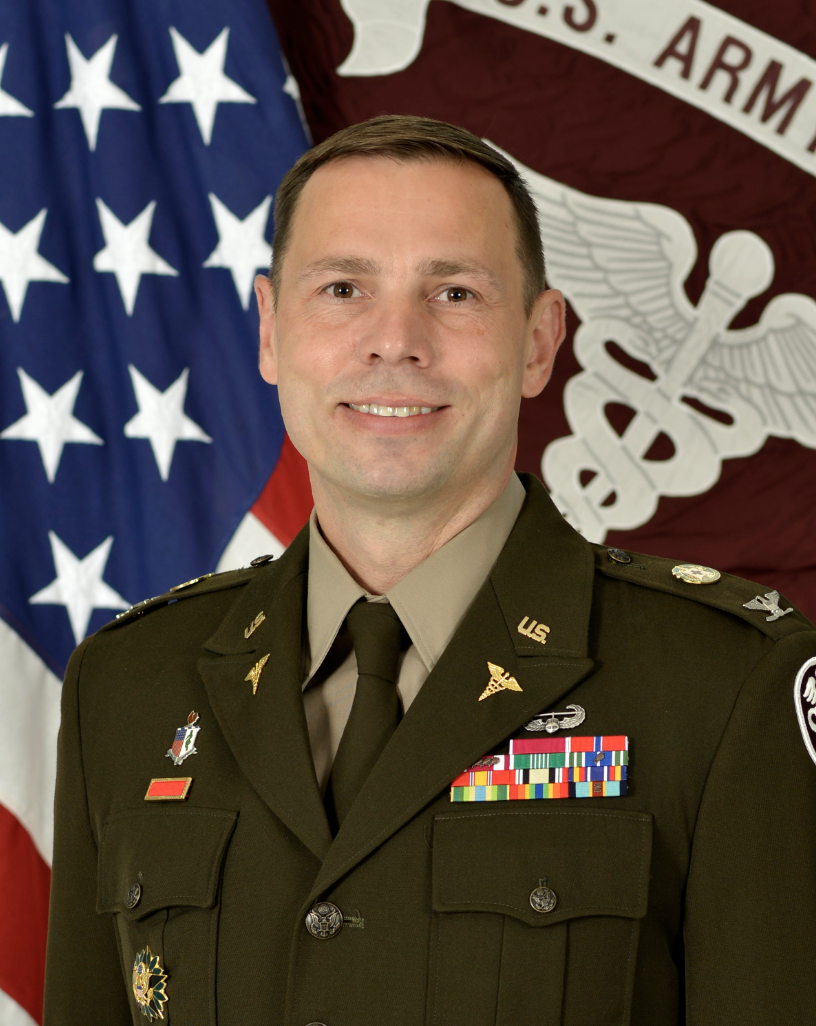}
\end{wrapfigure}

\textbf{Col.\ Dr.\ Preston Samuel L., DO, MS, FAPA} currently serves as Commander and CEO of Blanchfield Army Community Hospital, Fort Campbell, Kentucky. A dual board-certified physician in Family Medicine and Psychiatry, he graduated from the Lake Erie College of Osteopathic Medicine and completed residencies at Walter Reed Army Medical Center. His distinguished military career includes leadership roles such as Chief of Behavioral Health and Regional Director of Psychological Health across multiple Army commands. 
\\
\\




\begin{wrapfigure}{L}{0.2\textwidth}
\centering
\includegraphics[width=0.2\textwidth]{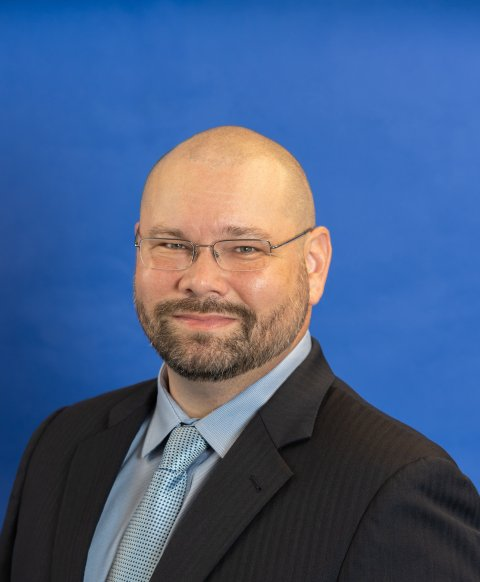}
\end{wrapfigure}

\textbf{Dr. Christopher K. Rhea} is the Associate Dean for Research \& Innovation in the College of Health Sciences at Old Dominion University. He is a recognized expert at the intersection of neuromotor control and advanced technology, using tools such as virtual reality and smartphones to address human health challenges like fall prevention in older adults, concussion assessment, and rehabilitation. His research has been supported by major agencies including the NIH, Department of Defense, US Navy, HRSA, and the Women’s Football Foundation, and he is known for building interdisciplinary teams to solve complex health problems.
\\
\\

\begin{wrapfigure}{L}{0.2\textwidth}
\centering
\includegraphics[width=0.2\textwidth]{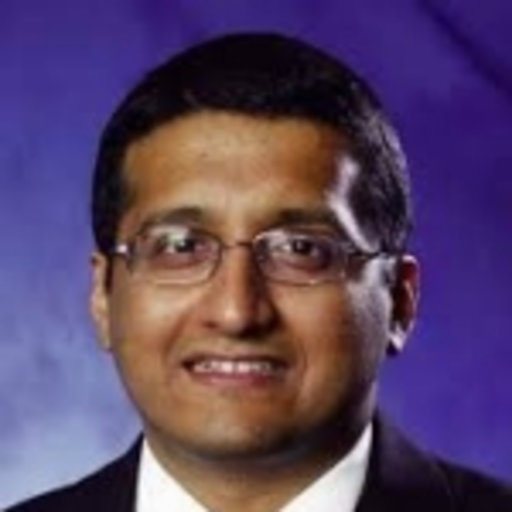}
\end{wrapfigure}

\textbf{Dr. Sachin Shetty} is an Associate Director in the Virginia Modeling, Analysis and Simulation Center at Old Dominion University and an Associate Professor with the Department of Computational Modeling and Simulation Engineering. Sachin Shetty received his PhD in Modeling and Simulation from the Old Dominion University in 2007. His research interests lie at the intersection of computer networking, network security and machine learning. Recently, he has been involved with developing cyber risk/resilience metrics for critical infrastructure and blockchain technologies for distributed system security. His laboratory has been supported by the National Science Foundation, Air Office of Scientific Research, Air Force Research Lab, Office of Naval Research, Department of Homeland Security, and Boeing.  He has published over 150 research articles in journals and conference proceedings and four books. He is the recipient of Commonwealth Cyber Initiative Research Fellow, Fulbright Specialist award, EPRI Cybersecurity Research Challenge award, DHS Scientific Leadership Award and has been inducted in Tennessee State University’s million-dollar club.
\\
\\

\end{document}